\pdfoutput=1

\documentclass[11pt]{article}

\usepackage[table]{xcolor}
\usepackage[preprint]{acl}

\usepackage{times}
\usepackage{latexsym}
\usepackage{todonotes}

\usepackage{inconsolata}

\usepackage[utf8]{inputenc} %
\usepackage[T1]{fontenc}    %
\usepackage{hyperref}       %
\usepackage{url}            %
\usepackage{booktabs}       %
\usepackage{amsfonts}       %
\usepackage{nicefrac}       %
\usepackage{microtype}      %
\usepackage{natbib}
\usepackage{graphicx}
\usepackage{subcaption}
\usepackage{fancyhdr}   %
\usepackage{multirow}
\usepackage{tabularx}
\usepackage{geometry}     %
\usepackage{comment}
\usepackage{amsmath, amssymb, amsthm}
\usepackage{xspace}
\usepackage{algorithmic}
\usepackage{algorithm}
\usepackage{stackengine}
\usepackage[breakable]{tcolorbox}
\usepackage{listings}
\usepackage{titlesec}
\titlespacing\paragraph{0pt}{0pt plus 1pt minus 1pt}{0pt plus 0pt minus 0pt}
\usepackage[nameinlink]{cleveref}
\crefname{figure}{Fig.}{Figs.}

\makeatletter
\renewcommand{\paragraph}{%
  \@startsection{paragraph}{4}%
  {\z@}{1ex \@plus 1ex \@minus .2ex}{-1em}%
  {\normalfont\normalsize\bfseries}%
}
\makeatother

\newcommand{\model}[1]{\texttt{#1}}
\newcommand{\dataset}[1]{\texttt{#1}}

\newif\ifshowcomments
\showcommentstrue  %

\makeatletter

\definecolor{lightblue}{RGB}{84, 151, 193}  

\newcommand{\UQm}{{\em UQ method}\@\xspace}
\newcommand{\UQms}{{\em UQ methods}\@\xspace}

\newcommand{\Corr}{{\em Correctness function}\@\xspace}
\newcommand{\corr}{{\em correctness function}\@\xspace}

\newcommand{\Corrs}{{\em Correctness functions}\@\xspace}
\newcommand{\corrs}{{\em correctness functions}\@\xspace}

\definecolor{lightpurple}{RGB}{192, 159, 189}  

\newcommand{\auroc}{{\em UQ performance metric}\@\xspace}
\newcommand{\aurocs}{{\em UQ performance metrics}\@\xspace}

\newcommand{\LMJ}{LM-as-a-judge\@\xspace}
\newcommand{\numUQest}{{ 8 }\@\xspace}

\makeatother

\lstdefinelanguage{Markdown}{
  morekeywords={[1]\#, \#\#, \#\#\#, \#\#\#\#, **, *, -, `}, 
  sensitive=false,
  morecomment=[l]{<!--}, %
  morecomment=[s]{<!--}{-->},
  morestring=[b]", %
}

\lstdefinestyle{mdstyle}{
  language=Markdown,
  basicstyle=\ttfamily\small,   %
  keywordstyle=\color{blue},     %
  commentstyle=\color{gray},     %
  stringstyle=\color{red},       %
  breaklines=true,               %
  showstringspaces=false         %
}

\title{Revisiting Uncertainty Quantification Evaluation in Language Models: \\ Spurious Interactions with Response Length Bias Results}

\author{%
  Andrea Santilli\thanks{Equal contribution}\thanks{Work done during an internship at Apple.}\\
  Sapienza University\\ of Rome\\
  \And
  Adam Goli\'{n}ski\footnotemark[1]\\
  Apple \\
  \And
  Michael Kirchhof \\
  Apple \\
  \And
  Federico Danieli \\
  Apple \\
  \AND
  Arno Blaas \\
  Apple \\
  \And
  Miao Xiong\footnotemark[2] \\
  National University\\ of Singapore \\
  \And
  Luca Zappella \\
  Apple \\
  \And
  Sinead Williamson\\
  Apple \\
}

\begin{document}

\maketitle

\begin{abstract}

Uncertainty Quantification (UQ) in Language Models (LMs) is key to improving their safety and reliability. Evaluations often use metrics like AUROC to assess how well \UQms (e.g., negative sequence probabilities) correlate with task \corrs (e.g., ROUGE-L). 
We show that {mutual biases}-when both \UQms and \corrs are biased by the same factors-{systematically distort evaluation}. First, we formally prove that any mutual bias non-randomly skews AUROC rankings, compromising benchmark integrity.
Second, we confirm this happens empirically by testing 7 widely used {\em correctness functions}, from lexical-based and embedding-based metrics to \LMJ approaches, across 4 datasets $\times$ 4 models $\times$ \numUQest \UQms. Our analysis shows that length biases in \corrs distort UQ assessments by interacting with length biases in \UQms.
We identify \LMJ methods as the least length-biased, offering a promising path for a fairer UQ evaluation.

\end{abstract}

\section{Introduction}
Language Models (LMs) excel at natural language generation %
but often produce factually incorrect outputs, or ``hallucinations'' \citep{guerreiro2023hallucinations,10.1145/3703155}.
These hallucinations are typically associated with high uncertainty about the correct output \citep{xiao2021hallucination}, leading to the emergence of {\em Uncertainty Quantification (UQ) methods} as a compelling approach to detect errors \cite{farquhar2024, baan2023uncertainty}.
A fundamental challenge in evaluating \UQms is the lack of ground truth uncertainty labels. Consequently, benchmarks commonly rely on \aurocs such as AUROC, assessing how effectively \UQms distinguish correct from incorrect outputs as determined by a \corr. Thus, the accuracy and reliability of UQ evaluations inherently depend on the quality of the correctness assessments.

In this paper, we critically analyze how errors and biases in \corrs impact \aurocs. First, we provide a formal analysis showing that: i) if errors in the \corr are random and independent from the \UQm, AUROC is noisy but unbiased; ii) conversely, if there exists a \textit{mutual bias}—i.e. if the \corr errors correlate systematically with the uncertainty scores—then AUROC rankings are inherently skewed. Our formal results demonstrate that \textbf{any mutual bias} introduces systematic distortions into AUROC evaluations, artificially advantaging certain methods and fundamentally undermining the reliability of benchmarks.

We confirm this happens empirically by benchmarking 7 widely-used \corrs, including lexical-based metrics (e.g., ROUGE metrics \citep{lin-2004-rouge}), embedding-based metrics (e.g., BERTScore \cite{Zhang2020BERTScore}), and \LMJ approaches \citep{lianmin2024judging} across 4 datasets $\times$ 4 models $\times$\numUQest \UQms.
We reveal two key issues: (i) the \corr choice significantly impacts UQ results and (ii) widely used lexical-based and embedding-based \corrs \cite{farquhar2024,fadeeva2023} introduce systematic biases that distort the perceived effectiveness of certain \UQms.

A human evaluation of 450 LM samples%
 reveals that this bias stems, at least in part, from the mutual dependence of certain \UQms and %
\corrs on the output length.
Building on this, we identify \corrs that mitigate bias by avoiding such confounding, finding \LMJ approaches best suited for UQ evaluation and most aligned with human judgment.
Overall, this study highlights widespread pitfalls in UQ evaluation and charts a path toward a more reliable evaluation protocol.

\section{Evaluating uncertainty}
In this section, we review common \UQms, \corrs and \aurocs. %

\subsection{\UQms}
\label{sec:uq_methods}
Given an input \(x\) to an LM, which generates an output sequence \(\hat{y}\), a \UQm estimates a measure of the model's uncertainty about \(\hat{y}\), denoted as \(\hat{g}(\hat{y}, x)\).
These methods can be broadly categorized into three types: (i) single-sample, (ii) multi-sample, and (iii) learned.
One simple single-sample approach is negative sequence probability, %
\begin{equation}
\label{eq:seq-prob}
\textstyle \hat{g}(\hat{y}, x) = -\hat{p}(\hat{y} | x) = -\prod_{i=1}^{L} \hat{p}(\hat{y}_i|\hat{y}_{<i}, x),
\end{equation}
where $L$ is the length of the generated answer and $\hat{p}$ the output probabilities assigned by the model. 
Note that in \autoref{eq:seq-prob}, $\hat{g}$ increases with \(L\).
Multiple-sample approaches derive uncertainty scores by sampling multiple responses for the same input \(x\) and measuring a metric (e.g., variance) across samples. Notably, several \UQms in this second group, like Na{\"i}ve Entropy and Semantic Entropy \cite{farquhar2024}, use \autoref{eq:seq-prob} to compute the probability of a sequence of generated tokens.
Lastly, learned methods train a binary classifier via supervised learning on a correctness-labeled dataset \cite{kadavath2022}.
We refer readers to \autoref{app:uq-methods} for more details on each family.

\begin{figure*}[th!]
    \centering
    \includegraphics[width=\textwidth]{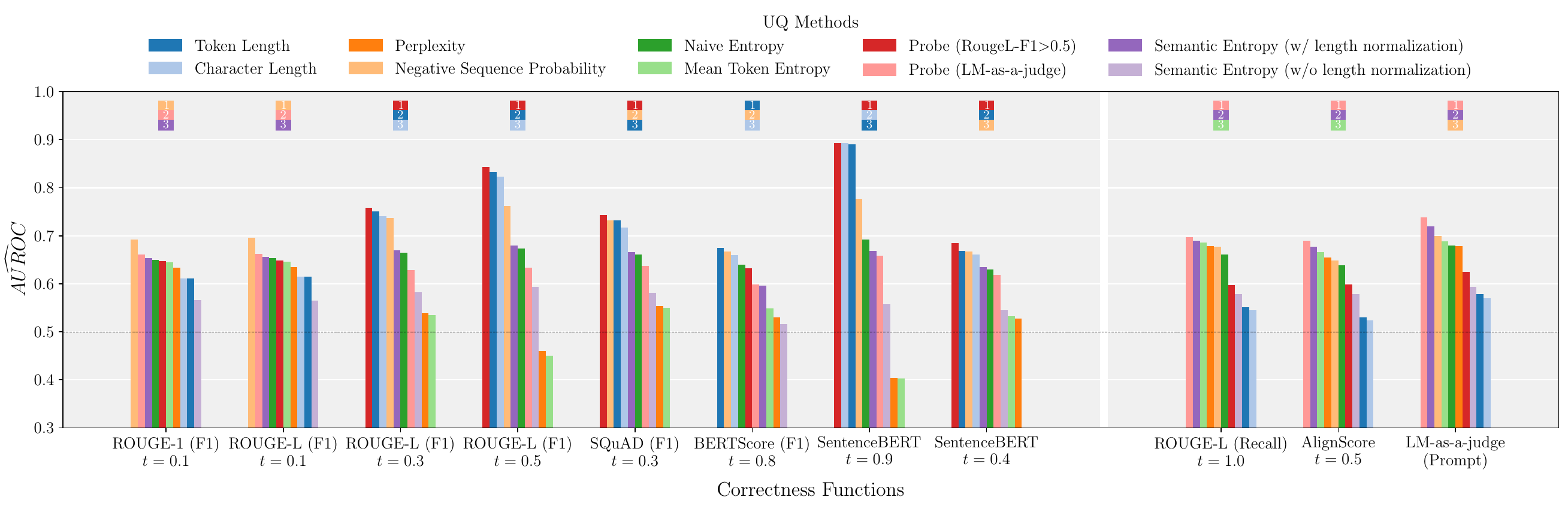}
    \caption{
    $\widehat{AUROC}$ of various \UQms across \corrs averaged over models and datasets. 
    The ranking of \UQms (top row) changes across \corrs, raising questions about which one to trust.
    }
    \label{fig:results_long_form}
\end{figure*}

\subsection{\Corrs}
\label{sec:correctness_metrics}
\Corrs \(\hat{h}(\hat{y}, x, y)\) compare a generated answer \(\hat{y}\) to a reference answer \(y\) to estimate a correctness score, and can be categorized as lexical-based, embedding-based, or \LMJ.

\paragraph{Lexical-based}
\corrs, such as SQuAD \citep{rajpurkar-etal-2016-squad} and ROUGE variants \citep{lin-2004-rouge}, are based on lexical overlap between $\hat{y}$ and $y$. %
While limitations of these metrics have been studied in areas like summarization and Question Answering (QA) \cite{guo-vosoughi-2023-length, chen-etal-2019-evaluating, cohan-goharian-2016-revisiting, fabbri2021summeval, reiter-belz-2009-investigation}, their impact on UQ evaluation remains largely unexplored.

\paragraph{Embedding-based}
\corrs, such as BERTScore \cite{Zhang2020BERTScore} and SentenceBERT cosine similarity \cite{reimers-gurevych-2019-sentence}, assess similarity by encoding both
$\hat{y}$ and $y$ %
using a language model, typically BERT-based.

\paragraph{\LMJ} 
\corrs evaluate correctness by using another LM to judge the accuracy of $\hat{y}$ against $y$. %
Examples %
include AlignScore \cite{zha-etal-2023-alignscore}, which uses a specifically trained LM, %
and prompt-based variants of \LMJ \cite{lianmin2024judging}.
\paragraph{}\hspace{-13.5pt}
\autoref{tab:correctness_metrics} summarizes common \corrs used in recent UQ papers. AUROC requires binary labels, so a certain threshold $t$ is typically applied to binarize continuous correctness scores. Some \corrs are inherently binary (e.g., \LMJ), and some \aurocs do not require binarization \cite{fadeeva2023}. This variety in UQ eval protocols raises questions about which combination to trust. \autoref{app:details_correctness_metrics} offers a broader view on each metric.

\begin{table}[t]
    \centering
    \resizebox{\columnwidth}{!}{%
    \begin{tabular}{lll}
        \toprule
        \textbf{\Corr} & \textbf{Used in UQ eval protocol}  & \textbf{Threshold $t$} \\
        \midrule
        ROUGE-1 (F1)& \citet{aichberger2024semantically} & $0.1-1.0$\\
        ROUGE-L (F1)& \citet{fadeeva2023, kuhn2023} & $0.5$\\ 
                 & \citet{duan-etal-2024-shifting, chen2024inside} & $0.5$\\
                 & \citet{qiu2024semantic} & $0.1-1.0$\\
                 & \citet{aichberger2024semantically} & $0.1-1.0$\\
        SQuAD (F1) & \citet{farquhar2024} & $0.3$\\
        \arrayrulecolor{gray!50} \cmidrule{1-3}
        BERTScore (F1)& \citet{fadeeva2023} & N/A \\
        SentenceBERT  & \citet{chen2024inside} & $0.9$\\
        \arrayrulecolor{gray!50} \cmidrule{1-3} \arrayrulecolor{black}
        AlignScore & \citet{vashurin2025benchmarkinguncertaintyquantificationmethods} & $0.5$ \\
        \LMJ (Prompt) & \citet{farquhar2024} & N/A \\ 
        \bottomrule
    \end{tabular}
    }
    \caption{\Corrs used in UQ evals.}
    \label{tab:correctness_metrics}
\end{table}

\subsection{\aurocs}
\label{sec:uq_perf_metrics}
The utility of \UQms is typically assessed using a \auroc that quantifies how well uncertainty estimates (\textsection \ref{sec:uq_methods}) correlate with correctness. %
Among the various \aurocs available in the literature \cite{malinin2020, fadeeva2023}, we focus on the Area Under the Receiver Operating Characteristic curve (AUROC)
due to its widespread use in UQ benchmarks \cite{farquhar2024, chen2024inside}.

Let $\hat{g}_i \equiv \hat{g}(\hat{y_i}, x_i)$ be the {uncertainty} score assigned by some \UQm to the \(i\)-th data sample, and let $h_i$ be a binary label denoting (ground truth) correctness of that data sample ($h_i = 1$ if correct, $h_i = 0$ if incorrect).
AUROC 
can be written %
as
\begin{equation}
\text{AUROC} = P\left( \hat{g}_i < \hat{g}_j
\mid h_i = 1, \; h_j = 0 \right),
\label{eq:auroc}
\end{equation}
i.e., the probability that a randomly chosen \emph{correct} data sample receives a lower %
uncertainty score than a randomly chosen \emph{incorrect} data sample.

\begin{figure}[th!]
\centering
\includegraphics[width=\columnwidth]{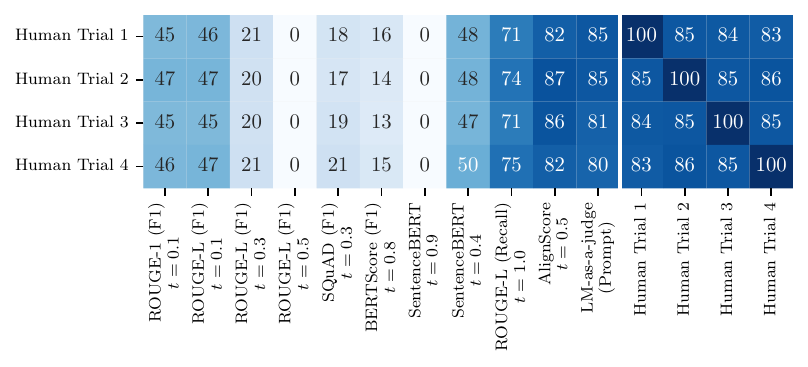}
\caption{
Cohen Kappa agreement rates between annotators and \corrs.
Per dataset: \cref{fig:app_agreement_datasets}.
}
\label{fig:agreement_main}
\end{figure}

\subsection{Mutual biases in \auroc}
\label{sec:uq_perf_error}
In practice, we estimate $h_i$ using a \corr $\hat{h}_i \equiv \hat{h}(\hat{y}_i, x_i, y_i)$ from \textsection \ref{sec:correctness_metrics}. This means that AUROC is not computed on ground-truth labels, but rather on a potentially biased surrogate:
\begin{equation}
\widehat{\text{AUROC}} =P\left( \hat{g}_i < \hat{g}_j
\mid \hat{h}_i = 1, \; \hat{h}_j = 0 \right).
\label{eq:auroc_est}
\end{equation}

\autoref{eq:auroc_est} highlights a key challenge: the measured performance is merely an \textit{estimate} of the true AUROC and is subject to \corr errors, which can interact with and distort the final outcome. Specifically, two scenarios may arise:\\
\textbf{i) Uncorrelated errors.} If errors in $\hat{h}$ are independent of $\hat{g}$, $\widehat{\text{AUROC}}$ is a noisy but \textit{unbiased} estimator of the real ${\text{AUROC}}$. While scores, in the worst case, regress toward the 0.5 random baseline, no \UQm is systematically favored or penalized.\\
\textbf{ii) Mutually biased errors.} If errors in $\hat{h}$ correlate with $\hat{g}$, the estimated performance of $\hat{g}$ will be \textit{systematically biased}. Depending on the direction of correlation, some \UQms may appear more or less effective than they truly are—leading to inflated or deflated evaluations and ultimately compromising the validity of performance comparisons.
These two scenarios are formally characterized and analyzed in \autoref{sec:uq-correctness-errors}, which provides a theoretical foundation for understanding how \corr errors propagate into \aurocs.

In practice, in \textsection \ref{sec:evaluating_uncertainty} we find that both \UQms and \corrs can exhibit biases over the answer length $L$, falling into scenario (ii). 
Understanding the impact of these biases is crucial to ensuring fair and reliable comparisons in UQ. %

\begin{figure*}[th!]
    \begin{minipage}[t]{0.2\textwidth} 
        \vspace{10pt} %
        \centering
        \includegraphics[width=\linewidth]{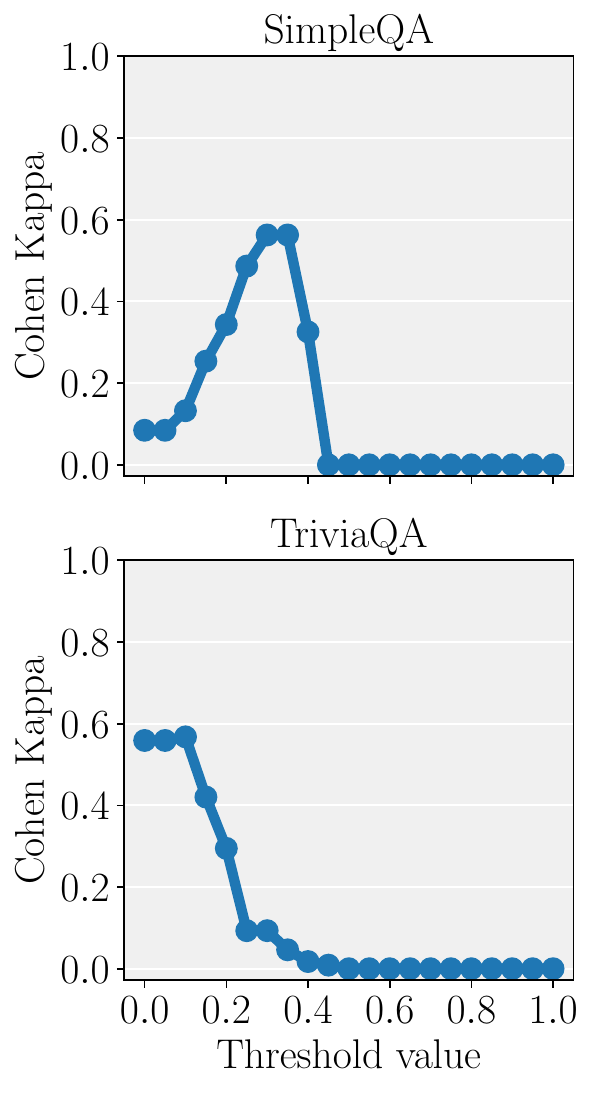}
        \captionof{figure}{
        Cohen Kappa score w.r.t.\ human annotators for Rouge-L (F1) against thresholds.
        }
        \label{fig:threshold}
    \end{minipage}%
    \hfill %
    \begin{minipage}[t]{0.79\textwidth} %
        \vspace{0pt} %
        \centering %

        \includegraphics[
          clip,
          width=\linewidth %
        ]{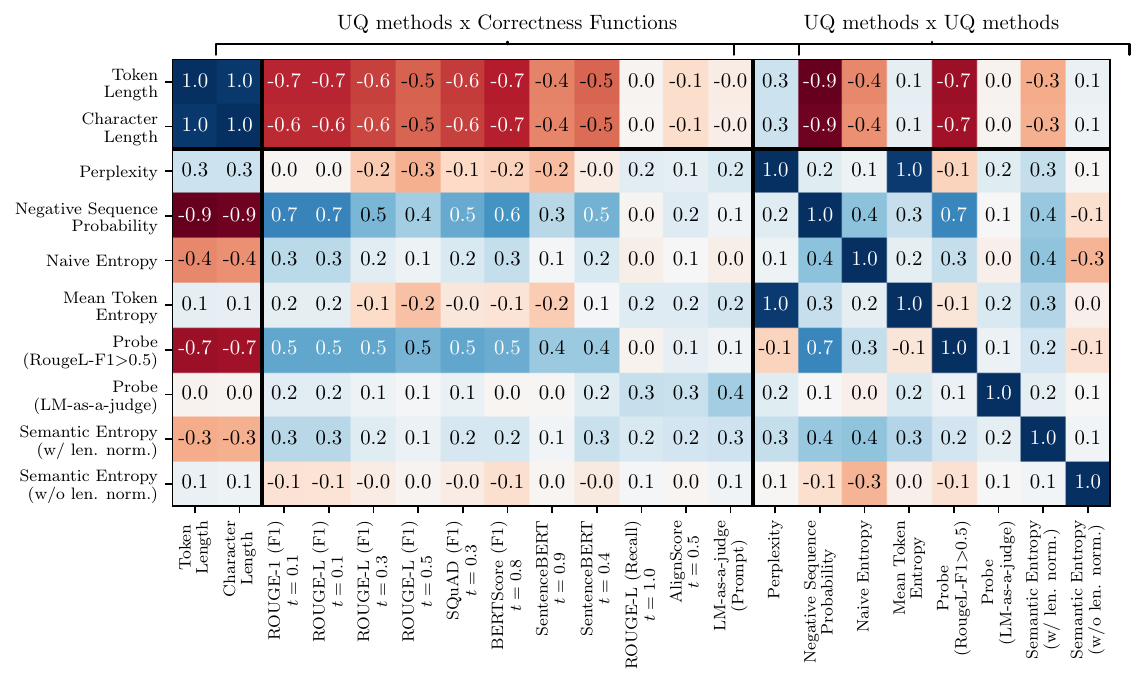}
        \captionof{figure}{Spearman’s rank correlation coefficients.}
        \label{fig:correlation_plot_main}
    \end{minipage}%
\end{figure*}

\section{Experiments}
\label{sec:evaluating_uncertainty}
In this section, we evaluate several \UQms following evaluation protocols in line with previous related works \citep{lin2024a,fadeeva2023,farquhar2024} while varying just the \corr. 
We consider generative QA tasks, as they are standard in UQ literature and their single-answer format simplifies correctness evaluation relative to more open-ended tasks like summarization.

\paragraph{Experimental setup.}
We evaluate the performance of \numUQest \UQms across 4 datasets, 4 models, and 7 \corrs. 
For details, see: 
\autoref{app:uq-methods} on \UQms;
\autoref{app:details_correctness_metrics} on \corrs;
\autoref{app:experimental-details} on models, datasets and prompts.

\subsection{Impact of the \corr on UQ}
\label{sec:ablating}

\cref{fig:results_long_form} illustrates the estimated performance of different \UQms when varying only the \corr. For each method, we report average $\widehat{\text{AUROC}}$ across datasets and models, focusing on the most commonly used \corrs from \autoref{tab:correctness_metrics}. \cref{fig:results_long_form} reveals that changing the \corr affects not only the estimated AUROC value but also the ranking of \UQms.
This, however, raises the question of which metric to trust and what is causing the disagreement.

\subsection{Evaluating \corrs for UQ}
\label{sec:evaluating-answer-evaluation-metrics}

The previous section shows how \corrs choices impact benchmarking conclusions, but it is unclear which function yields reliable UQ results. 
To investigate, we evaluate several \corrs against human annotations (four annotators per sample for 450 samples, see \autoref{app:detail_human_exp}).
The Cohen's Kappa \cite{cohen1960coefficient, artstein-poesio-2008-survey} values in \cref{fig:agreement_main}\footnote{Some approaches show no agreement at all due to poor thresholding choices.} %
show that \LMJ approaches (prompt-based and AlignScore) align best with human labelers, followed by ROUGE-L (Recall) with $t\!=\!1$. 

Revisiting \cref{fig:results_long_form}, we see that these three \corrs show more stable orderings, with some variability in AUROC magnitudes---consistent with expectations for small errors that are mostly uncorrelated with \UQms,
as per case (i) in \textsection \ref{sec:uq_perf_error}. 
Conversely, most previously-used lexical- and embedding-based \corrs poorly reflect human judgment.

\paragraph{Impact of threshold choice.}
A major source of error in lexical- and embedding-based \corrs stems from the thresholding strategy used to binarize scores for AUROC computation. As shown in Table \ref{tab:correctness_metrics}, prior work often applies standard thresholds or experiments with a small set of options. 
However, \cref{fig:agreement_main,fig:threshold} illustrate that metrics like ROUGE-L (F1) and SentenceBERT are highly sensitive to threshold choices, as assessed by the resulting agreement with humans. Poor thresholding can lead to degenerate outcomes—e.g., assigning nearly identical labels to all predictions—which drastically reduces alignment with human annotators. 
The issue is further exacerbated by the fact that optimal thresholds vary across tasks (\cref{fig:threshold,fig:app_threshold_human_agreement}) and are heavily influenced by response verbosity (\cref{fig:score_vs_length}), making it challenging to select a single effective threshold. 
In contrast, metrics such as ROUGE-L Recall, AlignScore, and LM-as-a-judge exhibit %
considerably less sensitivity to threshold selection, as shown in \cref{fig:app_corrs_vs_length} of the Appendix.

\subsection{Mutual bias in \corrs and \UQms}
\label{sec:interaction}

We concluded that \LMJ approaches achieve higher agreement with humans than other \corrs. 
However, this alone does not explain the shift in \UQm rankings observed in 
 \textsection \ref{sec:ablating}. If errors were random, no systematic effect would emerge, falling into case (i) of our error analysis (\textsection \ref{sec:uq_perf_error}). However, this is not the case: the relative performance of negative sequence probability, perplexity, and probes varies dramatically. %
 This is indicative of a spurious correlation between \UQms and errors in the \corrs.

\paragraph{Many \UQms are biased by length.}
Many \UQms explicitly or implicitly depend on the length of a response. In particular, negative sequence probability assigns higher uncertainty to longer responses, as each term in \autoref{eq:seq-prob} is $< \! 1$.
Other \UQms that incorporate \autoref{eq:seq-prob} in their computation (\textsection\ref{sec:uq_methods}), such as Naïve Entropy and Semantic Entropy, may also be impacted.
To investigate this %
relationship, we compute Spearman correlation %
between the scores from various \UQms and the length of generated answers (measured in tokens and characters). 
In \cref{fig:correlation_plot_main}, we see
that multiple estimators exhibit significant positive or negative correlations with length.

\begin{figure}[t]
    \begin{subfigure}[c]{0.495\columnwidth}
        \centering
        \includegraphics[width=\textwidth]{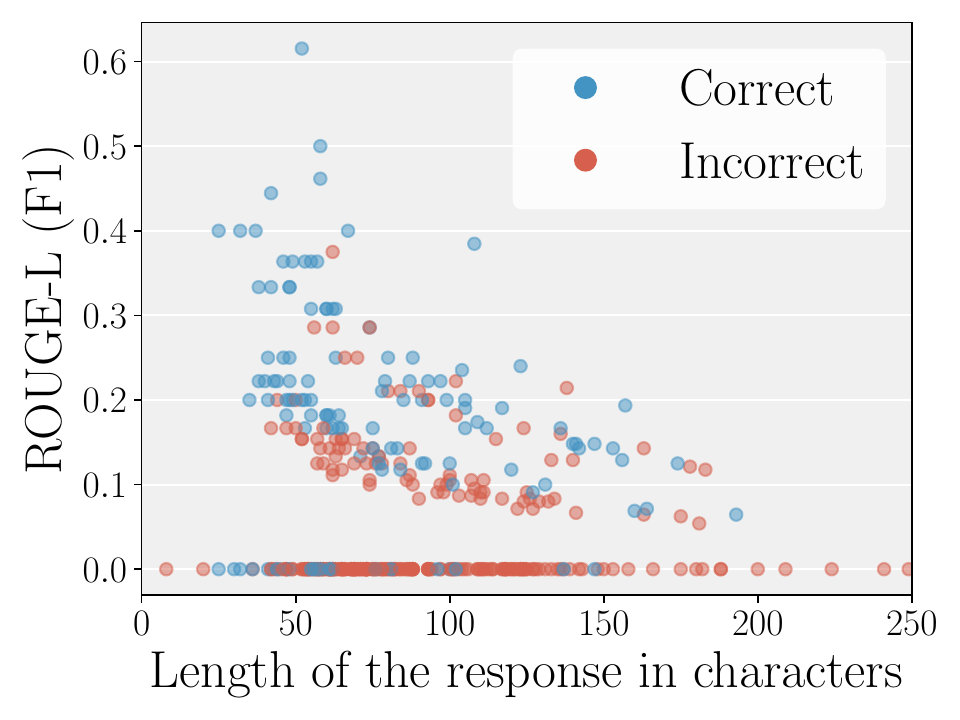}
        \caption{ROUGE-L (F1)}
        \label{fig:score_vs_length_response_rougel}
    \end{subfigure}
    \begin{subfigure}[c]{0.495\columnwidth}
        \centering
        \includegraphics[width=\textwidth]{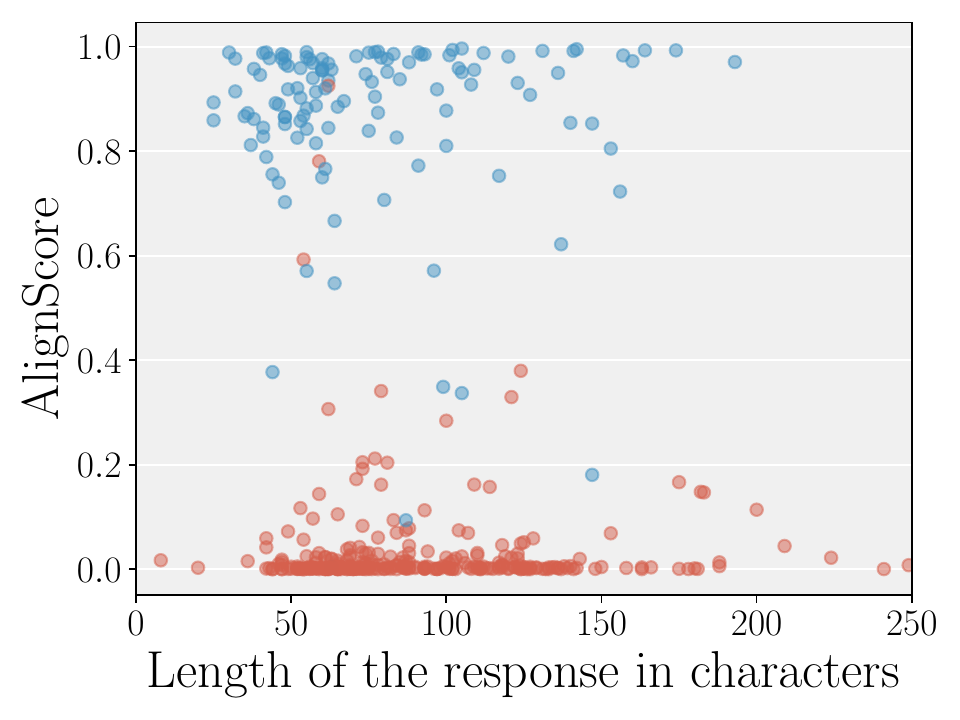}
        \caption{AlignScore}\label{fig:score_vs_length_response_alignscore}
    \end{subfigure}
    \caption{
        \Corr vs response length.
        The color indicates human correctness judgments.
        Results for other \corrs in the Appendix, \cref{fig:app_corrs_vs_length}.
    }
    \label{fig:score_vs_length}
\end{figure}

\paragraph{Many \corrs are biased by length}
Many \corrs are also known to exhibit length bias when assessing summaries \cite{guo-vosoughi-2023-length}. We demonstrate that this issue also affects QA.
In \cref{fig:score_vs_length}, we analyze the relationship between response length and \corr output, showing correctness values for responses where all annotators agree on the label.
The ROUGE-L (F1) score  is highly dependent on response length, favoring shorter sentences and making threshold selection challenging. In contrast, AlignScore is length-independent and clearly separates correct and incorrect samples.
\autoref{app:additional_results} presents similar findings for other \corrs.

\paragraph{Spurious interaction.}

Mutual correlation between \UQms and \corrs on answer length can systematically inflate or deflate \aurocs (\autoref{sec:uq-correctness-errors}). This effect is evident in \cref{fig:results_long_form}, where length-based baselines (token and character length—blue bars) perform competitively on lexical- and embedding-based \corrs but rank last under \LMJ metrics. This may also explain discrepancies in prior works, such as the inflated ranking of negative sequence probability in \citet{fadeeva2023}.
We recommend using \LMJ where possible, as its lower error is less likely to impact \aurocs.
While ROUGE-L (Recall) is inherently independent of the generated answer's length, %
it offers lower correlation with human judgment, leading to noisier AUROC estimates and offering a higher likelihood of additional confounding variables. 
For example, it is vulnerable to exploitation by models that produce multiple off-target answers alongside the correct one.

\section{Beyond Length Bias}
In this paper, we argue that \textbf{any biases}—not just length—present \textbf{simultaneously} in both the \UQm and the \corr induce a spurious correlation that systematically biases the \auroc (AUROC). Crucially, this effect does not merely introduce random noise (i.e., increased variance) to the AUROC estimate; rather, it leads to consistent bias, producing misleading results that can {artificially favor} certain \UQms over others. We support this claim through both analytical derivation (\autoref{sec:uq-correctness-errors}) and empirical evidence (\cref{fig:results_long_form}).
Importantly, this is a general result. While we use length bias as a running example—because it is is the most severe and already impacting benchmarks—the underlying issue extends to \textbf{any confounding variable that correlates with both \UQms \corrs}. 
Identifying such confounders, which are less obvious than length, is inherently difficult, which makes their presence particularly dangerous for evaluation protocols.
For instance, in settings that combine LM-as-a-judge approaches with verbalized uncertainty methods (e.g., \citet{huang-etal-2024-calibrating, 10.5555/3692070.3692180, yang2024logu}), one might hypothesize less obvious and harder-to-detect sources of confounding like vocabulary used or writing style of the response \cite{feuer2025styleoutweighssubstancefailure}.
Our goal is not to enumerate all possible biases but to establish the existence of a broader class of systematic evaluation failures—of which length bias is a concrete and empirically validated case. Identifying and mitigating other such biases remains an important direction for future work.

\section{Conclusion}
We prove that \aurocs are systematically biased when the \UQm and \corr share a confounder. 
Empirically, we identify response length as a concrete instance of such mutual bias which is affecting existing benchmarks and undermines their reliability.

Our results highlight that lexical- and embedding-based \corrs, commonly used in prior work, frequently introduce these distortions. In contrast, \LMJ approaches exhibit greater robustness and stronger alignment with human judgments, making them a more reliable choice for UQ evaluation. That said, we recommend validating any \LMJ setup against human annotations before applying it to new tasks or datasets~\cite{bavaresco2024llmsinsteadhumanjudges}.

\section*{Limitations}

In this work, we critically examine the role of the \corr in the evaluation of \UQms using \aurocs. While our analysis sheds light on biases introduced by \corrs, certain limitations remain.  

Our analysis is focused on the context of QA, as it is a standard task in UQ literature and provides well-defined single-answer questions, making the definition of a \corr easier compared to open-ended tasks like machine translation and summarization where even objective human judgment of correctness is difficult. However, previous work suggests that the length bias of errors in \corrs is not unique to the QA setting \citep{guo-vosoughi-2023-length}, suggesting that \aurocs will face similar issues in such tasks.

Our recommendation is to use \LMJ as a potential \corr. While using another LM to judge correctness has demonstrated advantages \cite{lianmin2024judging}, it also comes with known limitations \cite{wang-etal-2024-large-language-models-fair, chen-etal-2024-humans}. The reliability of the correctness assessment may vary depending on the choice of the judging LM and the prompt formulation. More concerningly, if the same LM is used as both the \corr and as part of the \UQm, we are likely to have correlations between the \LMJ's errors and the \UQm, which could inflate the \UQm's performance---although this is mitigated by the relatively low frequency of such errors. Additionally, while our analysis on QA is based on widely used QA datasets, we do not know whether the same LM judge and prompts would generalize effectively to other tasks and datasets. Ideally, an LM judge should be rigorously evaluated against human annotators before being employed in new tasks and datasets \cite{bavaresco2024llmsinsteadhumanjudges}. Furthermore, \LMJ introduces significant computational overhead compared to traditional \corrs, making it less practical for resource-constrained applications.  

Finally, our study identifies response length as a confounding factor in UQ benchmarking, but other latent variables may also influence \UQms and \corrs in subtle ways, as discussed in \autoref{sec:uq-correctness-errors}. A deeper understanding of these biases is crucial for refining UQ evaluation protocols and ensuring more reliable assessments of model uncertainty.

\section*{Acknowledgements}
We would like to thank Xavier Suau, Miguel Sarabia, Pau Rodríguez, and Eugene Ndiaye for their feedback on an earlier draft of this manuscript.

\bibliography{references_adam,references}

\clearpage
\appendix

\section{Details of \UQms}
\label{app:uq-methods}

There are several methods for generating uncertainty estimates that help assess the in-correctness of LM outputs. These approaches can be broadly classified into three main categories:
\textbf{1) Single-sample methods}: methods that require a single forward pass from the model and that generally use directly the logits and probability distributions over the vocabulary space provided as output from the model; 
\textbf{2) Multiple-sample methods}: methods that, given a prompt $x$, sample multiple possible outputs for the same prompt and compute an uncertainty score based on these outputs; 
\textbf{3) Learned methods}: usually probes or small networks directly trained to predict the accuracy of the model given the prompt and the answer.

We denote with $x$ the sequence of tokens corresponding to the prompt. This usually includes the instruction prompt (e.g., "Answer the following question") together with the question and additional context. The $L$ generated tokens are indicated as $\hat{y}_i$. Additionally, a superscript $\hat{y}^{(s)}$ is used for multiple-sample methods to indicate the $s$-th sample (out of $S_{\text{UQ}}$ samples) sampled for a given prompt. $\hat{p}(\cdot)$ denotes the probability assigned by the model.

\paragraph{Single-sample methods.} 

Single-sample methods estimate the uncertainty score using the logits that the models output. These logits are usually computed on the greedy decoded output or on a low-temperature sample decoded from the model given the prompt $x$.

\emph{Negative Sequence Probability.} \quad
Sequence probability computes the cumulative probability of the sequence. This can be used as an uncertainty score by flipping the sign and considering $-\hat{p}(\hat{y} | x)$. 
When evaluated on the greedy decoding samples, this method is sometimes referred to in the literature as \textit{Maximum Sequence Probability} (MSP) \cite{fadeeva2023, vashurin2025benchmarkinguncertaintyquantificationmethods}.
\citet{aichberger2024semantically} investigate the difference between the performance of MSP estimated using greedy decoding and estimated using the Beam Search decoding algorithm, which yields sequences with higher likelihood.
\begin{equation}
\label{eq:seq-prob-app}
\hat{p}(\hat{y} | x) = \prod_{i=1}^{L} \hat{p}(\hat{y}_i|\hat{y}_{<i}, x).
\end{equation}

\emph{Perplexity.} \quad
Perplexity computes the uncertainty score via the exponential of the mean token likelihood \cite{4767370}. Compared to sequence probability, perplexity normalizes the underlying probability by the number of the generated tokens,
\begin{equation}
\exp\left( - \frac{1}{L} \sum_{i=1}^{L} \log \hat{p}(\hat{y}_i|\hat{y}_{<i}, x)\right).
\end{equation}

\emph{Mean Token Entropy.} \quad 
Mean token entropy \cite{fomicheva2020unsupervised,malinin2020} computes the mean of the per-token entropies over the vocabulary distribution,
\begin{equation}
\mathcal{H}_T(\hat{y}, {x}) = \frac{1}{L} \sum_{i=1}^{L} \mathcal{H}\left[ \hat{p}(\hat{y}_i|\hat{y}_{<i}, x) \right].
\end{equation}

\paragraph{Multiple-Sample methods.} 
Multiple-sample methods compute an uncertainty score by sampling $S_{\text{UQ}}$ times for a single prompt.
Since it is accessing (more of) the full probability distribution, this class of methods should provide better uncertainty scores than single-sample methods, albeit at the expense of an increased computational cost at inference time. The exact number of samples $S_{\text{UQ}}$ is a hyperparameter that usually depends on the specific \UQm.

\emph{Naive Entropy.}  \quad
Naive Entropy computes the entropy over the different generated samples. The sequence probability of each generation is computed using the chain rule of probability, like in the {Sequence Probability} method,
\begin{equation}
\label{eq:naive-entropy}
-\sum_{s=0}^{S_{\text{UQ}}} \hat{p}(\hat{y}^{(s)} | x) \log \hat{p}(\hat{y}^{(s)} | x).
\end{equation}

\emph{Semantic Entropy.} \quad 
Semantic entropy computes the entropy over the different semantic clusters $C$ of the generated samples \cite{farquhar2024}.
Semantic clusters are generated using a Natural Language Inference (NLI) model, which evaluates bidirectional entailment between pairs of answers in 
$S_{\text{UQ}}$. This process group answers with equivalent meanings into clusters $c^{(i)}$. 
Each cluster probability $\hat{p}(c^{(i)})$ is computed by summing the {Sequence Probabilities} of the unique generations that fall into that cluster \cite{farquhar2024}. The probability of each generated sequence is computed using \autoref{eq:seq-prob-app}, either directly for Semantic Entropy (without length normalization), or normalized by the sequence token-length $L$ for length-normalized Semantic Entropy.
\begin{equation}
\label{eq:semantic-entropy}
SE(x) = -\sum_{i=1}^C \hat{p}(c^{(i)}|x) \log \hat{p}(c^{(i)}|x).
\end{equation}

\paragraph{Learned methods.} 
Learned methods leverage the model's internal activations or its entire architecture to train additional networks or classifiers that predict the correctness of the answer.

\emph{Probes} are the most common form of learned method. The most prominent variety of probe is \emph{P(IK)}, also known as P(I Know) \cite{kadavath2022}, which finetunes the entire model to predicts a binary score whether the model can answer the question correctly or not. This is accomplished by attaching a classifier to the embedding of the final token in the last layer. The training set is collected by labeling some generations from the model with the task \corr. In this paper, we follow the implementation of \citep{farquhar2024, kapoor2024large} that does not train the full model but just a logistic regression classifier on top of the representation of the final answer token as in \cite{chen2024inside}.
We trained probes using two different \corrs: \LMJ (using \texttt{Qwen/Qwen2.5-72B-Instruct}), and ROUGE-L (Recall) with a 0.5 threshold.  When reporting results, we specify the \corr used to label the dataset and train the probe in round brackets—for example, \texttt{Probe(\LMJ)}, where \LMJ denotes the judging model. 
Probes are trained until convergence on each training dataset with L-BFGS and a tolerance value of $0.0001$ and maximum number of optimization iterations of $10000$.

\section{Details of \Corrs}
\label{app:details_correctness_metrics}
In this section, we describe in detail the \corrs used in our experiments. Many of these metrics return a continuous score, which is binarized for calculating AUROC; we detail the thresholds $t$ used in this binarization below.

\subsection{Lexical-based}
Lexical-based metrics assess similarity by measuring lexical overlap between the generated sentence and the ground truth. These metrics are among the most widely used due to their low computational cost and long-standing history in QA evaluation.

\paragraph{}
It is important to note that these metrics were originally used to evaluate QA in \textit{trained systems}, where the output distribution of generated sequences has been aligned with the expected distribution of ground-truth answers in the dataset. However, in common LM zero-shot evaluation settings, this alignment is no longer guaranteed. Consequently, these metrics may fail to accurately assess correctness, requiring careful consideration when applying them. While techniques like incorporating few-shot examples, as demonstrated by \citet{farquhar2024}, can mitigate this issue to some extent, 
this does not fully address the fundamental limitations of lexical-based metrics.

\paragraph{ROUGE-L.} ROUGE-L measures the longest common subsequence (LCS) between the generated response and the reference answer, allowing for non-contiguous matches \cite{lin-2004-rouge}. In UQ the F1-score (\emph{ROUGE-L (F1)}) of this metric is typically used, balancing precision and recall \citep{fadeeva2023, kuhn2023, duan-etal-2024-shifting, chen2024inside, qiu2024semantic, aichberger2024semantically}. 
\emph{ROUGE-L (Precision)} measures the ratio of the longest common subsequence (LCS) length to the number of unigrams in the generated answer. 
\emph{ROUGE-L (Recall)} measures the ratio of the LCS length to the number of unigrams in the reference answer.
ROUGE-L (F1) is the harmonic mean of ROUGE-L precision and recall.
It is important to note that ROUGE-L recall is not affected by the length of the generated answer, whereas precision and F1 metrics are influenced by it.
In the experiments of this paper, we consider \emph{ROUGE-L (F1)} and \emph{ROUGE-L (Recall)} variants, with both metrics computed using the Python package \texttt{rouge\_scorer}.
Both ROUGE-L variants return continuous scores; where a binary score is used, we consider thresholds $t\in\{0.1, 0.3, 0.5\}$ for ROUGE-L (F1), and $t=1.0$ for ROUGE-L (Recall).

\paragraph{ROUGE-1.} ROUGE-1 measures the unigrams overlap between the generated response and the ground truth \cite{lin-2004-rouge}. ROUGE-1 captures similarity based on single-tokens overlap. This metric has been widely used in QA evaluations and in UQ benchmarks in \citet{aichberger2024semantically}. As in \citet{aichberger2024semantically}, we use the F1 variant (\emph{ROUGE-1 (F1)}). In the experiments of this paper, the metric has been computed using the Python package \texttt{rouge\_scorer}. ROUGE-1 (F1) returns continuous scores; where a binary score is used, we use a threshold $t=0.1$.

\paragraph{SQuAD.} This metric has been introduced in \citet{rajpurkar-etal-2016-squad} to measure the performance of systems trained on the homonymous dataset. The metric computes the F1 score based on word overlap between the prediction and ground truth, treating them as unordered bags of tokens, selecting the highest F1 among multiple references per question, and averaging across all questions. This metric has been used to evaluate correctness for UQ benchmarks in \citet{farquhar2024}. To compute the metric we used the implementation from \citet{von-werra-etal-2022-evaluate}. SQuAD returns continuous scores; where a binary score is used, we use a threshold $t=0.3$.

\subsection{Embedding-based}
Embedding-based metrics assess similarity by encoding both the ground truth and generated text using a neural model, typically BERT-based. The goal is to measure semantic similarity rather than surface-level overlap.

\paragraph{BERTScore.}
BERTScore \cite{Zhang2020BERTScore} evaluates generated answers by embedding both the generated text and the ground truth using a BERT pretrained model. It then computes the pairwise cosine similarity between tokens. For each token in the generated text, the highest similarity score with any token in the reference text is selected. Finally, precision, recall, and F1-score are calculated, with the F1-score commonly used in UQ to balance precision and recall.
This metric has been used to evaluate correctness for UQ benchmarks in \citet{fadeeva2023}.
In the experiments of this paper, the metric has been computed using the Python package \texttt{bert\_score} as in \citet{fadeeva2023}. BERTScore returns continuous scores; where a binary score is used, we use a threshold $t=0.8$, which we empirically found to yield the highest agreement with human raters in \cref{fig:agreement_main}.

\paragraph{SentenceBERT Similarity.}
A SentenceBERT model \cite{reimers-gurevych-2019-sentence} is used to encode both the generated answer and the ground truth answer. Specifically, following \citet{chen2024inside}, we use \textit{nli-roberta-large}\footnote{https://huggingface.co/sentence-transformers/nli-roberta-large}. The cosine similarity is then calculated between the ground truth and generated answer embeddings.  
This metric has been used to evaluate correctness for UQ benchmarks in  \citet{chen2024inside}. SentenceBERT returns continuous scores; where a binary score is used, we use a threshold $t\in\{0.4, 0.9\}$.

\subsection{\LMJ methods}
\LMJ metrics evaluate correctness by using another LM to judge the accuracy of a generated answer against the reference answer from the dataset. The evaluating LM may be specifically trained for this task or not.

\paragraph{AlignScore.}
AlignScore is a metric designed to evaluate the factual consistency of generated text with respect to a ground truth answer \cite{zha-etal-2023-alignscore}. It employs a RoBERTa model \citep{liu2019roberta} trained to assess the alignment between two text pieces, determining how well the generated content corresponds to the source information. The training process integrates data from several NLP tasks—natural language inference, question answering, paraphrasing, fact verification, information retrieval, semantic similarity, and summarization—resulting in a model trained specifically to evaluate correctness.
This metric has been used to evaluate correctness for UQ benchmarks in  \citet{vashurin2025benchmarkinguncertaintyquantificationmethods}. AlignScore returns continuous scores; where a binary score is used, we use a threshold $t=0.5$.

\paragraph{\LMJ (Prompt).}
\LMJ \cite{lianmin2024judging} encompasses a set of approaches that rely on a large language model to provide a human-like assessment of generated content by comparing it against a reference answer. Generally, different prompting strategies can be applied to guide the evaluation process. 
In our experiments, we used the same prompt as \citet{farquhar2024} with \texttt{Qwen/Qwen2.5-72B-Instruct} as the judging model. \LMJ returns binary scores, so no thresholds are used.

\newpage
\section{Impact of Correlated and Uncorrelated Errors in the \Corr on AUROC Estimation}
\label{sec:uq-correctness-errors}

In this section, we analyze the impact of correlated or uncorrelated errors in the \corr on AUROC estimation. We note that a similar analysis is performed in \citet{ielanskyi2025}, a concurrent work that explores impact of bias and variance of \corrs on AUROC estimation.

\noindent
\paragraph{}
Let \(\hat{g}_i \equiv \hat{g}(\hat{y}_i, x_i) \in \mathbb{R}\) be the uncertainty (UQ) score assigned to the answer \(\hat{y}_i\) given the question \(x_i\). We use \(h_i \equiv h(\hat{y}_i, x_i) \in \{0,1\}\) to denote the \emph{ground-truth correctness} of \(\hat{y}_i\), i.e., \(h_i = 1\) if the answer is correct and \(0\) otherwise. The \emph{estimated correctness} under some \corr \(\hat{h}\) is \(\hat{h}_i \equiv \hat{h}(\hat{y}_i, x_i, y_i) \in \{0,1\}\), possibly using a reference answer \(y_i\).

We define:
\[
    \text{TPR} \;=\; P\bigl(h=1 \mid \hat{h}=1\bigr), 
    \quad
    \text{FPR} \;=\; 1 - \text{TPR},
\]
\[
    \text{TNR} \;=\; P\bigl(h=0 \mid \hat{h}=0\bigr), 
    \quad
    \text{FNR} \;=\; 1 - \text{TNR}.
\]
The \emph{true AUROC} of \(\hat{g}\), based on ground-truth labels, is
\[
    \text{AUROC}(\hat{g}) \;=\; 
    P\bigl(\hat{g}_i < \hat{g}_j \;\mid\; h_i = 1,\; h_j=0\bigr).
\]
When correctness is measured by \(\hat{h}\), we obtain 
\[
    \widehat{\text{AUROC}}(\hat{g}) \;=\; 
    P\bigl(\hat{g}_i < \hat{g}_j \;\mid\; \hat{h}_i=1,\; \hat{h}_j=0\bigr).
\]
We additionally assume \(\displaystyle P(\hat{g}_i = \hat{g}_j) = 0\) for \(i \neq j\), implying 
\(\displaystyle \text{AUROC}(\hat{g}) = 1 - \text{AUROC}(-\hat{g}).\)

\paragraph{Expanding \(\mathbf{\widehat{AUROC}}\).}
We rewrite \(\widehat{\text{AUROC}}(\hat{g})\) by conditioning on both the true labels \((h_i, h_j)\) and the estimated labels \((\hat{h}_i, \hat{h}_j)\):

\medskip
\resizebox{\columnwidth}{!}{%
\(
\begin{aligned}
&\widehat{\text{AUROC}}(\hat{g})\\
    &= \sum_{a,b \in \{0,1\}}
       P\bigl(g_i < g_j \mid \hat{h}_i = 1,\; \hat{h}_j = 0,\; h_i=a,\; h_j=b\bigr) \\
    &\quad \times P\bigl(h_i=a \mid \hat{h}_i=1\bigr) \;\cdot\; P\bigl(h_j=b \mid \hat{h}_j=0\bigr)
\end{aligned}
\)
}%
\medskip

We discuss below two cases: 
(\emph{i})~when \(\hat{h}\)'s errors are \emph{independent} of \(\hat{g}\), and
(\emph{ii})~when they are \emph{correlated}.

\subsection{Case 1: Independent Errors}
\label{subsec:uncorrelated}
\noindent
\noindent
\textbf{Analysis}\;
\(\hat{h}_i \;\perp\!\!\!\perp\; \hat{g}_i \;\mid\; h_i\). In this setting, we have
\begin{equation}
    \begin{aligned}
        &\widehat{\text{AUROC}}(\hat{g})\\ &= \sum_{a,b\in \{0,1\}} P(g_i<g_j|h_i=a, h_j=b) \\
        &\quad \cdot P(h_i=a|\hat{h}_i=1)\cdot P(h_j=b|\hat{h}_j=0)\\
        \\
        &= P(g_i<g_j|h_i=0, h_j=0)\cdot \text{FPR}\cdot \text{TNR}\\
        &\quad + P(g_i<g_j|h_i=1, h_j=1)\cdot \text{TPR}\cdot \text{FNR}\\
        &\quad + P(g_i<g_j|h_i=1, h_j=0)\cdot \text{TPR}\cdot\text{TNR}\\
        &\quad + P(g_i<g_j | h_i=0, h_j=1)\cdot \text{FPR}\cdot\text{FNR}\\
        \\
        &= 0.5\cdot \text{FPR}\cdot \text{TNR}+ 0.5\cdot \text{TPR}\cdot \text{FNR}\\
        &\quad + P(g_i<g_j|h_i=1, h_j=0)\cdot \text{TPR}\cdot\text{TNR}\\
        &\quad + P(g_i<g_j | h_i=0, h_j=1)\cdot \text{FPR}\cdot\text{FNR}\\
        \\
        &= 0.5\cdot \text{FPR}\cdot \text{TNR}+ 0.5\cdot \text{TPR}\cdot \text{FNR}\\
        &\quad + \text{AUROC}(\hat{g}) \cdot \text{TPR}\cdot\text{TNR}\\
        &\quad + (1-\text{AUROC}(\hat{g}))\cdot \text{FPR}\cdot\text{FNR}.
    \end{aligned}
    \label{eq:auroc_uncorr}
\end{equation}

\bigskip

All terms \(\{\text{TPR},\text{TNR},\text{FPR},\text{FNR}\}\) in  \autoref{eq:auroc_uncorr}  are \emph{constant} properties of the \corr \(\hat{h}\), and do not depend on the \UQm $\hat{g}$. Hence \(\widehat{\text{AUROC}}(\hat{g})\) becomes a ``noisy'' version of the true \(\text{AUROC}(\hat{g})\), biased toward \(0.5\).

\noindent
\paragraph{Implications.}
In this uncorrelated setting, the bias introduced by \(\hat{h}\) \emph{does not} depend on the UQ metric \(\hat{g}\). Consequently, while the estimated AUROC values will be inaccurate, the ranking of UQ methods by \(\widehat{\text{AUROC}}(\hat{g})\) will, in expectation, match the ranking by \(\text{AUROC}(\hat{g})\), provided $\text{TPR}\cdot \text{TNR} > \text{FPR}\cdot \text{FNR}$. In practice, finite-sample effects can lead to variance, but with appropriate sample sizes, comparisons based on \(\widehat{\text{AUROC}}(\hat{g})\) remain valid.

\subsection{Case 2: Correlated errors}
\smallskip
\noindent
\textbf{Analysis}\;
\(\hat{h}_i \;\not\!\perp\!\!\!\perp\; \hat{g}_i \;\mid\; h_i\). In this case,

\begin{equation*}
    \begin{aligned}
P&(g_i<g_j|h_i=a, h_j=b, \hat{h}_i=1, \hat{h}_j=0) \\
&\neq P(g_i<g_j|h_i=a, h_j=b).
\end{aligned}
\end{equation*}
In particular, if the \corr's errors are negatively correlated with our \UQm (i.e., the more confident the \UQm is on a task, the more likely it is to be erroneously marked as correct), then we have
\begin{equation*}
\begin{aligned}P&(g_i<g_j|h_i=a, h_j=b, \hat{h}_i=1, \hat{h}_j=0) \\&> P(g_i<g_j|h_i=a, h_j=b),
\end{aligned}
\end{equation*}

for all values of $a$ and $b$, with the magnitude of the difference increasing with the magnitude of the correlation. This implies that $\widehat{\text{AUROC}}(\hat{g}) > \text{AUROC}(\hat{g})$. Similarly, if the errors are positively correlated with UQ metric, then we have $\widehat{\text{AUROC}}(\hat{g}) < \text{AUROC}(\hat{g})$.

This indicates that we will over-estimate the true AUROC if we have negatively correlated errors, and under-estimate if we have positively correlated errors. This is problematic because it introduces errors in AUROC that \textit{do} depend on the \UQm under consideration, leading to potential reordering of metrics.

\paragraph{Sources of Correlation.}
Since, in general, the \UQm does not depend on the output of the \corr or vice versa, any correlation between the \UQm, and errors in the \corr, must be due to information in $\hat{y}$ and/or $x$ that a) introduces systemic errors in the \corr, and b) is used by the \UQm. In this paper, we look at length as such a confounding variable, but it is not the only possible option. For example, the use of less frequently occurring words in $\hat{y}$ might lead to both an increase in uncertainty scores due to unfamiliar language, and an increase in the probability of erroneously marking an answer as incorrect due to reduced lexical overlap with the reference answer. We leave the exploration of additional confounders as future work.

\section{Experimental Details}
\label{app:experimental-details}

The datasets considered are \dataset{TriviaQA} \cite{joshi-etal-2017-triviaqa}, \dataset{SQuAD} \cite{rajpurkar-etal-2016-squad}, \dataset{NQ-Open} \cite{lee-etal-2019-latent}, and \dataset{SimpleQA} \cite{wei2024measuringshortformfactualitylarge}. The models considered are \model{Falcon-7B} \cite{penedo2023refinedweb}, \model{Qwen2.5-7B} \cite{qwen2025qwen25technicalreport}, and two versions of \model{Mistral-7B} \cite{jiang2023mistral}.

Our evaluation setup closely follows the methodology proposed by \citet{farquhar2024}\footnote{\href{https://github.com/jlko/semantic_uncertainty/}{https://github.com/jlko/semantic\_uncertainty/}}.
To obtain model responses, we employed the same prompt as in \citet{farquhar2024} for the long-form setting, instructing the model as follows:

\begin{quote}
\textit{Answer the following question in a single brief but complete sentence.}
\end{quote}

Responses are sampled using greedy decoding.
Similarly, for the \LMJ evaluation, we adhered to the same prompt of \citet{farquhar2024} and used the model \texttt{Qwen/Qwen2.5-72B-Instruct} as the judging model.
For our experiments, we employed the following models from the Hugging Face Hub \texttt{mistralai/Mistral-7B-Instruct-v0.1}, \texttt{mistralai/Mistral-7B-Instruct-v0.3}, \texttt{Qwen/Qwen2.5-7B-Instruct}, and \texttt{tiiuae/falcon-7b-instruct}.
The datasets used in our evaluation consist primarily of closed-book QA datasets, with the exception of \dataset{SQuAD} which is an open-book dataset.
Specifically, for \dataset{SQuAD} we incorporated the available context as part of the prompt.
In semantic clustering-based methods (Semantic Entropy), we employed DeBERTa as our Natural Language Inference (NLI) module as in \citet{farquhar2024}.

\begin{figure*}[t]
    \begin{subfigure}[c]{0.33\textwidth}
        \centering
        \includegraphics[width=\textwidth]{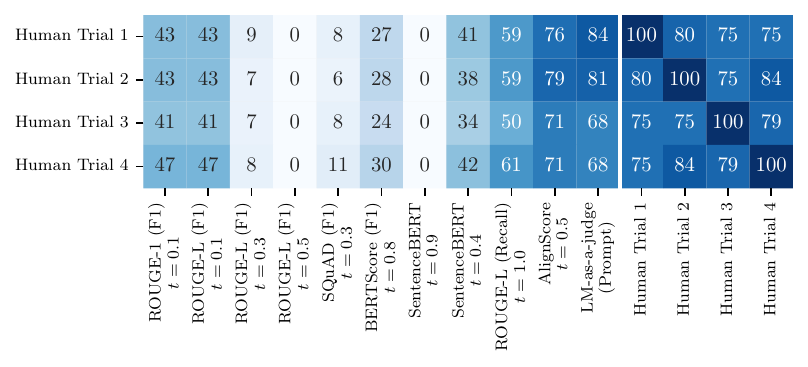}
        \caption{TriviaQA}
    \end{subfigure}
    \begin{subfigure}[c]{0.33\textwidth}
        \centering
        \includegraphics[width=\textwidth]{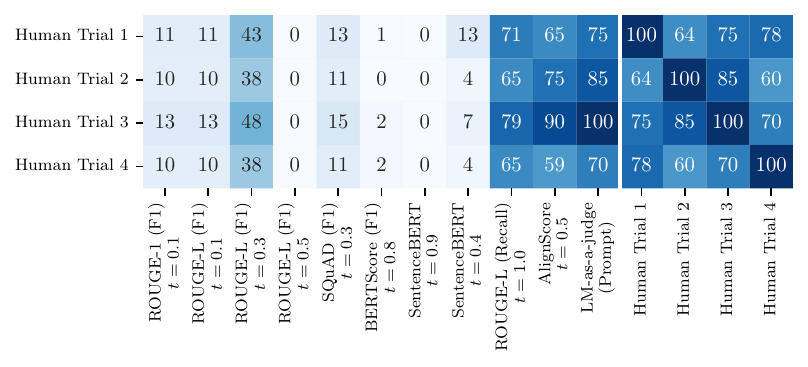}
        \caption{SimpleQA}
    \end{subfigure}
    \hfill
    \begin{subfigure}[c]{0.33\textwidth}
        \centering
        \includegraphics[width=\textwidth]{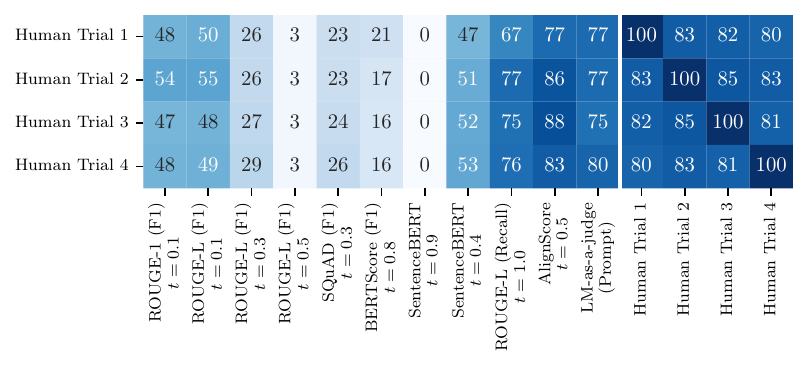}
        \caption{NQ}
    \end{subfigure}
    \caption{
    Cohen Kappa agreement rates between human annotators and \corrs.
    }
    \label{fig:app_agreement_datasets}
\end{figure*}

\section{Details on Human Annotation process}
\label{app:detail_human_exp}

We used an internal crowdsourcing platform to gather annotations. The raters were fluent English speakers and were compensated at or above the minimum wage. We randomly sampled 450 data points from \dataset{TriviaQA} \cite{joshi-etal-2017-triviaqa}, \dataset{NQ-Open} \cite{lee-etal-2019-latent}, and \dataset{SimpleQA} \cite{wei2024measuringshortformfactualitylarge}. We then generated  answers using \texttt{Qwen2.5-7B-Instruct}, with with greedy decoding.
We excluded \dataset{SQuAD} from the human annotation process to avoid incorporating the additional context into the annotation prompt, thereby streamlining and accelerating the annotation process. 
We then tasked human annotators to evaluate the correctness, collecting four annotations per data point.
Below, we present the annotation guidelines provided to each annotator. Each dataset included in the guidelines two manually labeled examples.

In \cref{fig:agreement_main} and \cref{fig:app_agreement_datasets}, we present the Cohen's Kappa agreement rates among human annotators. The first figure reports agreement computed across all 450 data points, while the second breaks down the agreement rates for each individual dataset (150 data points each).
For clearer visualization, \cref{fig:score_vs_length} and \cref{fig:app_corrs_vs_length} display a uniformly sampled subset of 150 data points from the full set of 450. These points represent correctness values for responses where all annotators agreed on the label.

\begin{tcolorbox}[colback=gray!10, colframe=black, title=Annotation Guidelines, breakable]
You will be provided with three pieces of text:

\begin{itemize}
    \item \textbf{Question}: A question previously proposed to a system or user. You are not asked to answer this question; it has already been answered (see Proposed Answer).
    \item \textbf{Reference Answer}: The authoritative or "gold standard" answer provided for the Question.
    \item \textbf{Proposed Answer}: The response given by a user for the Question that needs to be evaluated.
\end{itemize}

Your task is to determine whether the \textbf{Proposed Answer} is \textbf{equivalent} to the \textbf{Reference Answer} in the context of the \textbf{Question}.

\section*{Evaluation Criteria}

\begin{itemize}
    \item \textbf{Focus on Equivalence}: Assess whether the Proposed Answer conveys the same meaning as the Reference Answer, regardless of additional details or alternative phrasings.
    \item \textbf{Ignore Personal Knowledge}: Do not rely on your own knowledge or conduct external research. Base your judgment solely on the given text.
    \item \textbf{Exact Matching is Not Required}: The Proposed Answer does not need to be identical to the Reference Answer, but it must preserve the core meaning.
    \item \textbf{Context Matters}: Ensure that the Proposed Answer is relevant to the Question and correctly aligns with the Reference Answer's meaning.
\end{itemize}

\section*{Rating Scale}

Choose one of the following ratings for each evaluation:

\begin{itemize}
    \item \textbf{Equivalent}: The Proposed Answer conveys the same meaning as the Reference Answer.
    \item \textbf{Not Equivalent}: The Proposed Answer does not convey the same meaning, either due to missing essential information, contradictions, or incorrect interpretation.
\end{itemize}

\section*{Additional Notes}

\begin{itemize}
    \item \textbf{More Detail vs. Different Information}: Extra information is acceptable as long as the meaning remains the same. However, if the additional details introduce contradictions, the Proposed Answer should be marked \textbf{Not Equivalent}.
    \item \textbf{Paraphrasing is Allowed}: The wording of the Proposed Answer does not need to match exactly, but the core meaning must remain intact.
    \item \textbf{Avoid Assumptions}: Do not infer additional meaning beyond what is explicitly stated.
\end{itemize}

\section*{Examples}

\subsection*{Example 1: Equivalent}
<Follows one annotated equivalent example from the dataset>

\subsection*{Example 2: Not Equivalent}
<Follows one annotated not equivalent example from the dataset>
\end{tcolorbox}

\section{Additional Results}
\label{app:additional_results}

We present here supplementary results that were excluded from the main paper.

\cref{fig:app_agreement_datasets} presents the Cohen's Kappa agreement rate with human annotators, broken down by dataset. \LMJ approaches demonstrate stronger alignment with human judgments, whereas lexical-based and embedding-based \corrs are highly sensitive to the selection of an appropriate threshold.

\begin{figure*}
    \begin{subfigure}[c]{0.49\textwidth}
        \centering
        \includegraphics[width=\textwidth]{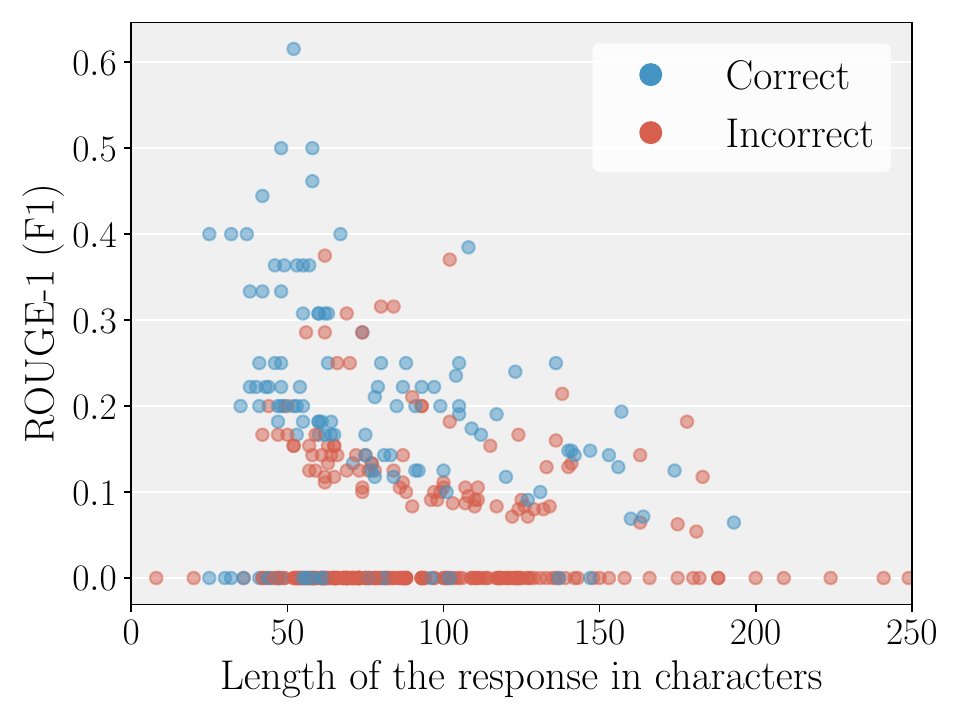}
    \end{subfigure}
    \hfill
    \begin{subfigure}[c]{0.49\textwidth}
        \centering
        \includegraphics[width=\textwidth]{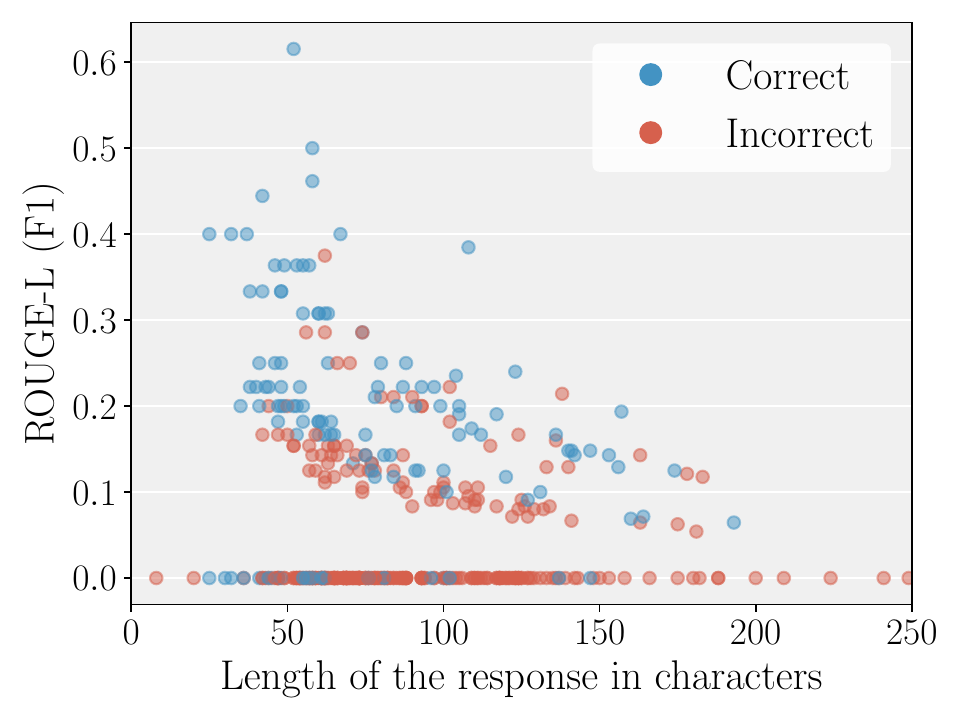}
    \end{subfigure}
    \begin{subfigure}[c]{0.49\textwidth}
        \centering
        \includegraphics[width=\textwidth]{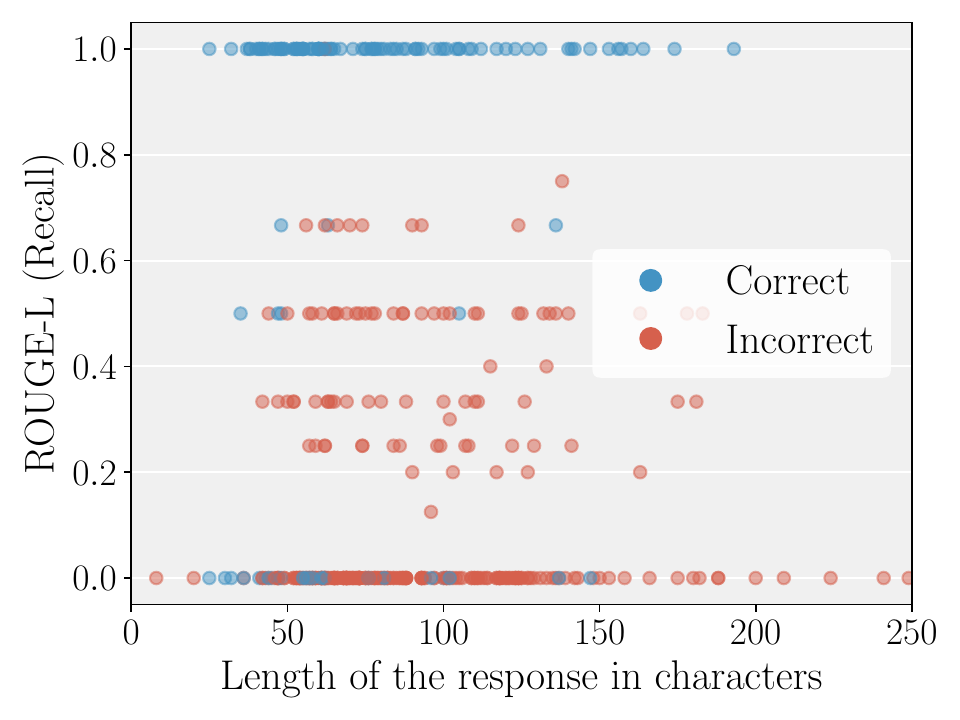}
    \end{subfigure}
    \hfill
    \begin{subfigure}[c]{0.49\textwidth}
        \centering
        \includegraphics[width=\textwidth]{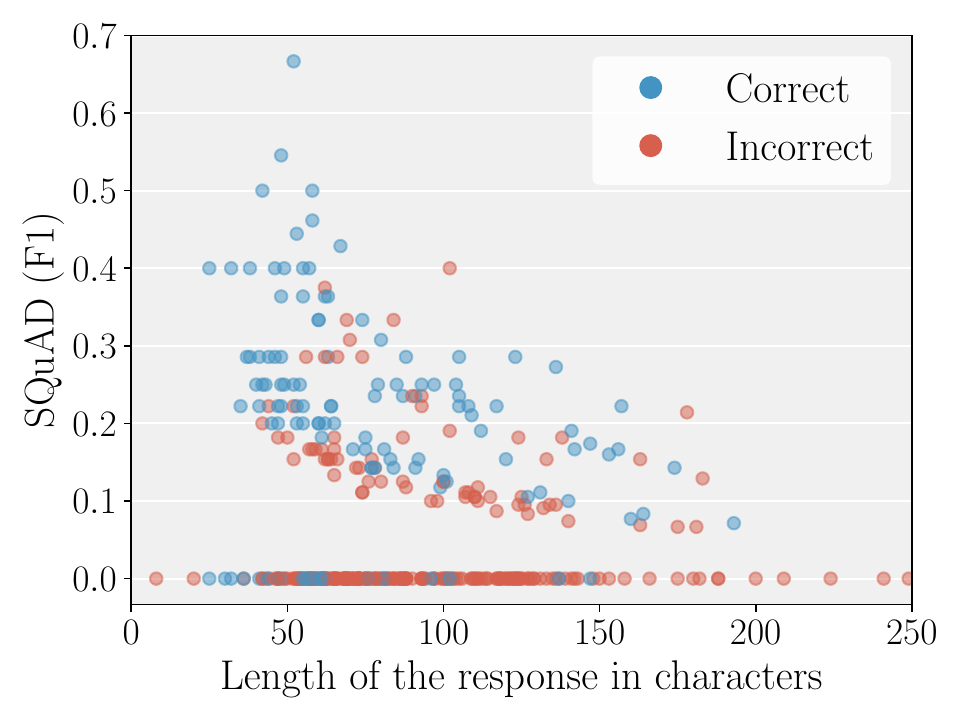}
    \end{subfigure}
    \begin{subfigure}[c]{0.49\textwidth}
        \centering
        \includegraphics[width=\textwidth]{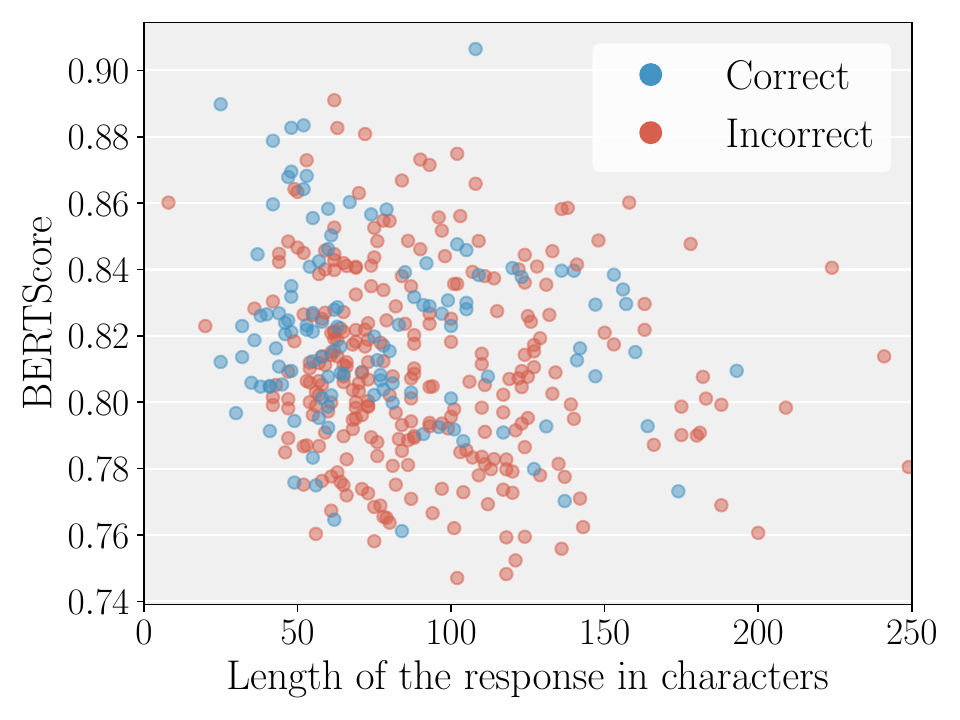}
    \end{subfigure}
    \hfill
    \begin{subfigure}[c]{0.49\textwidth}
        \centering
        \includegraphics[width=\textwidth]{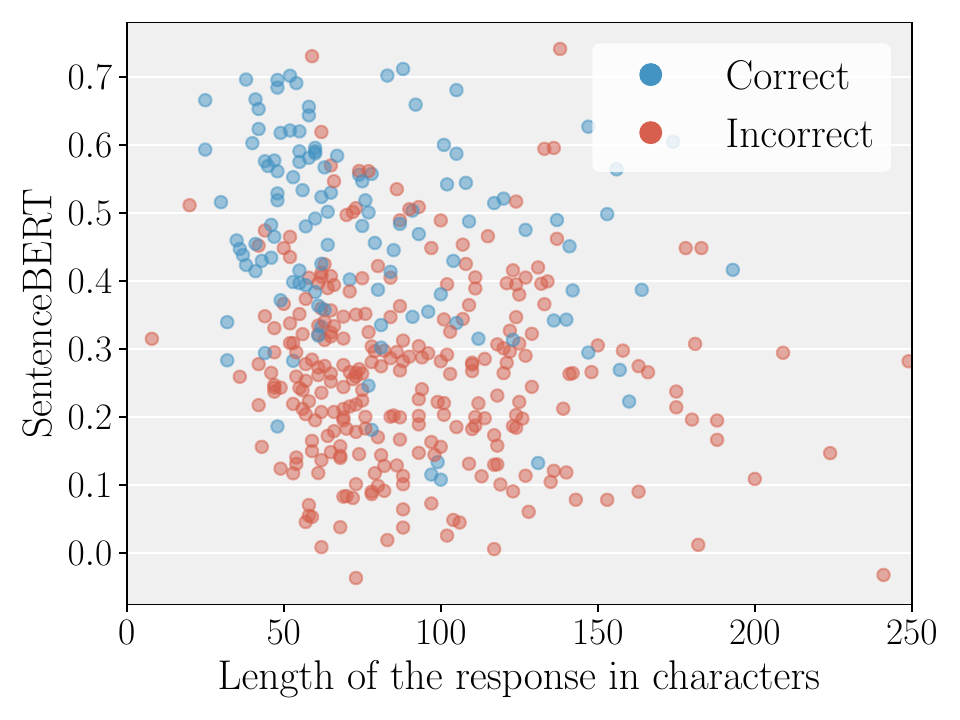}
    \end{subfigure}
    \begin{subfigure}[c]{0.49\textwidth}
        \centering
        \includegraphics[width=\textwidth]{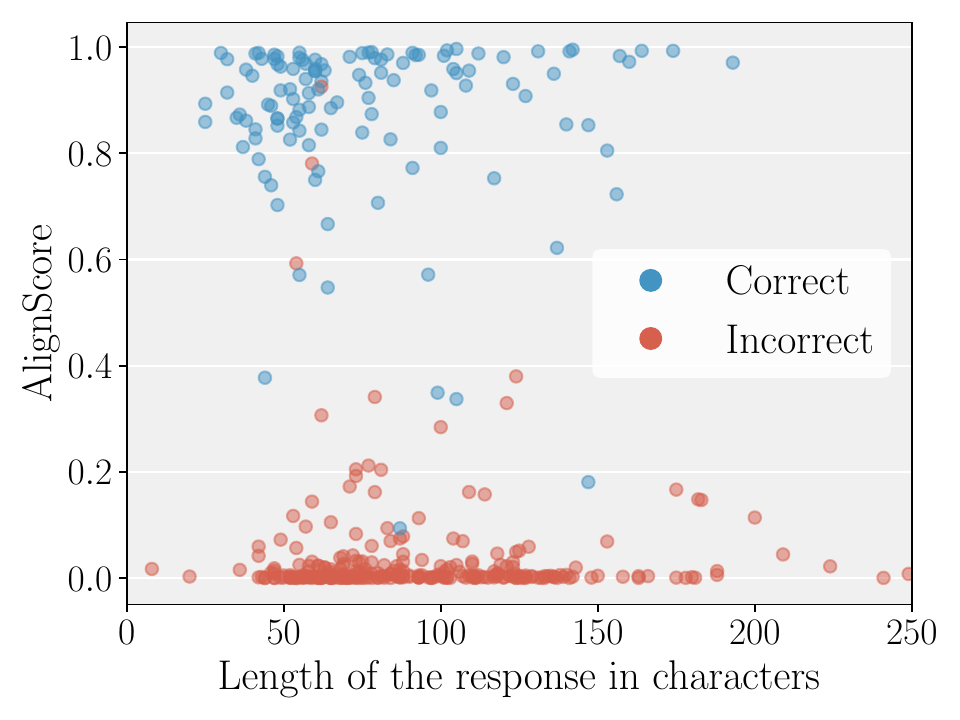}
    \end{subfigure}
    \hfill
    \begin{subfigure}[c]{0.49\textwidth}
        \centering
        \includegraphics[width=\textwidth]{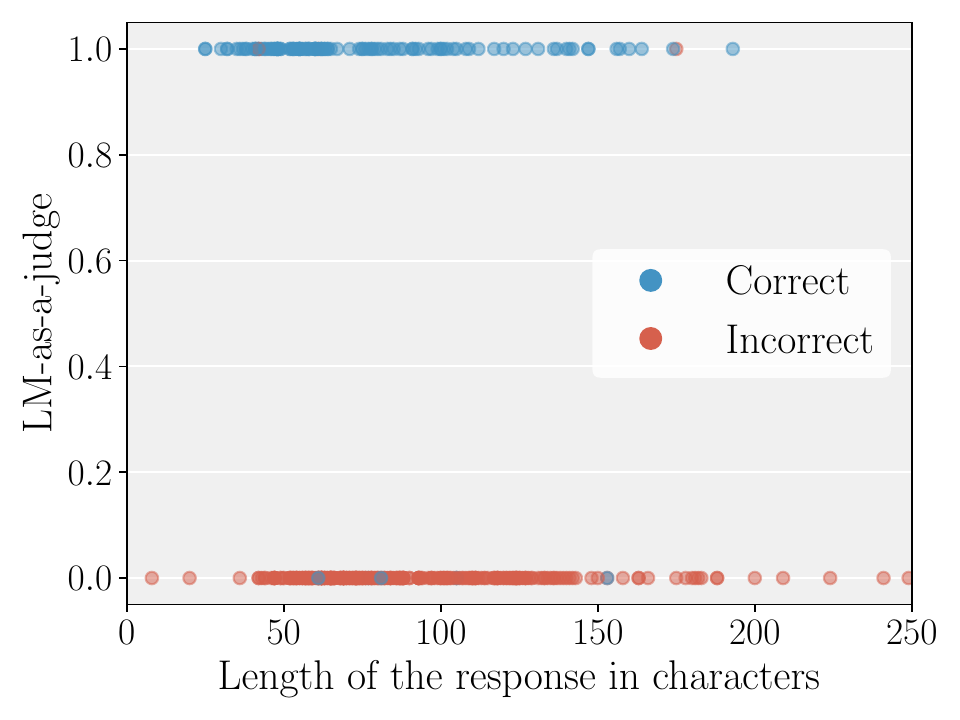}
    \end{subfigure}
    \caption{
    \Corr vs response length. Color indicates human correctness judgments.
    }
    \label{fig:app_corrs_vs_length}
\end{figure*}

\cref{fig:app_corrs_vs_length} illustrates the score assigned by \corrs as a function of the generated answer's length. 
Among the evaluated approaches, \LMJ methods (AlignScore and \LMJ \texttt{Qwen/Qwen2.5-72B-Instruct}) appear to be the only robust ones that remain invariant to length while effectively distinguishing between correct and incorrect samples without requiring threshold tuning.

\begin{figure*}[t]
  \centering
  
  \hspace{0.05\textwidth}%
  \begin{minipage}{0.28\textwidth}\centering
    \hspace{8pt}NQ
  \end{minipage}%
  \begin{minipage}{0.28\textwidth}\centering
    \hspace{8pt}SimpleQA
  \end{minipage}%
  \begin{minipage}{0.28\textwidth}\centering
    \hspace{8pt}TriviaQA
  \end{minipage}

  \begin{minipage}{0.05\textwidth}
    \centering
    \rotatebox{90}{Rouge-L (F1)}
  \end{minipage}%
  \begin{minipage}{0.28\textwidth}\centering
    \includegraphics[width=0.9\textwidth]{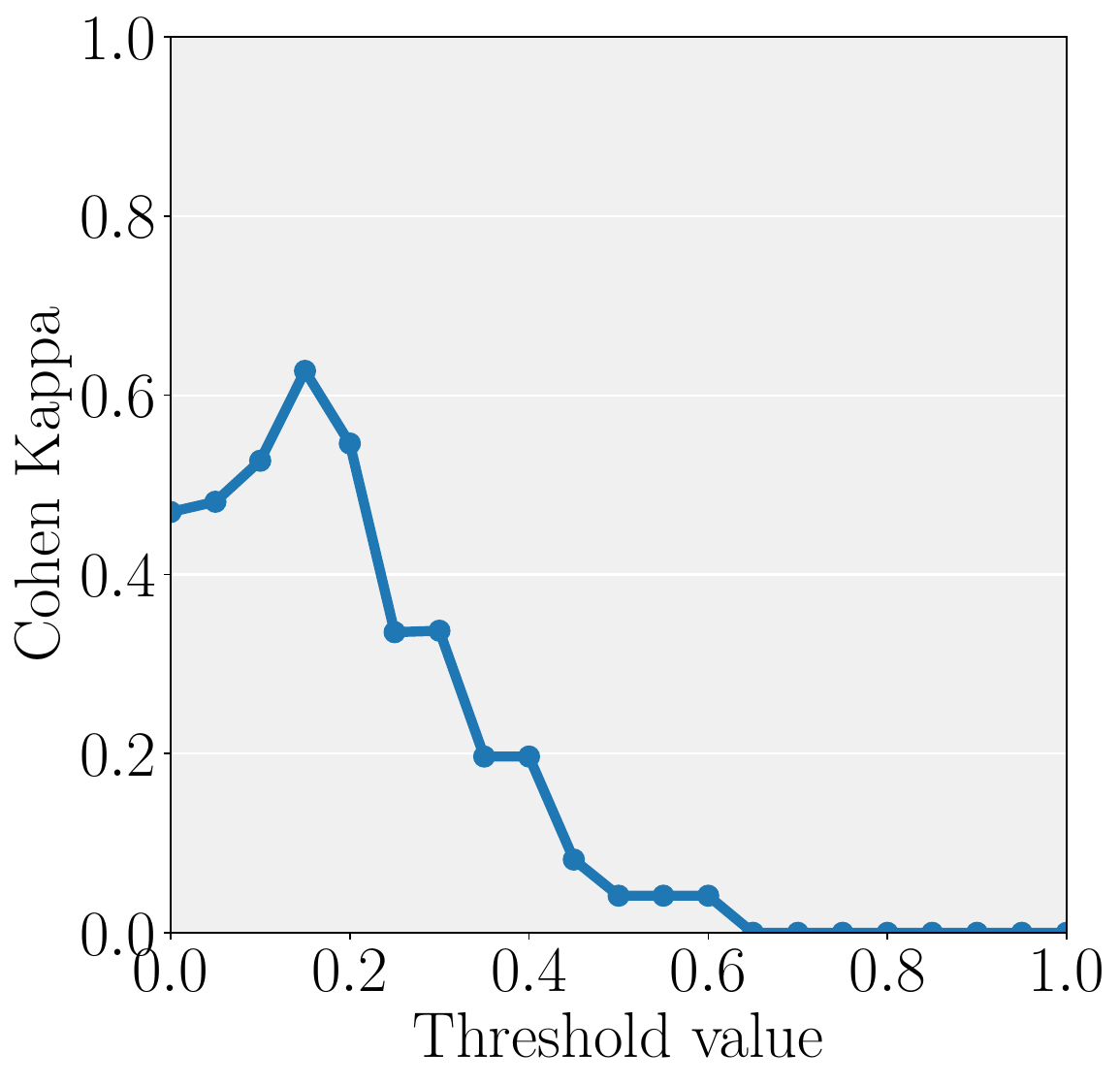}
  \end{minipage}%
  \begin{minipage}{0.28\textwidth}\centering
    \includegraphics[width=0.9\textwidth]{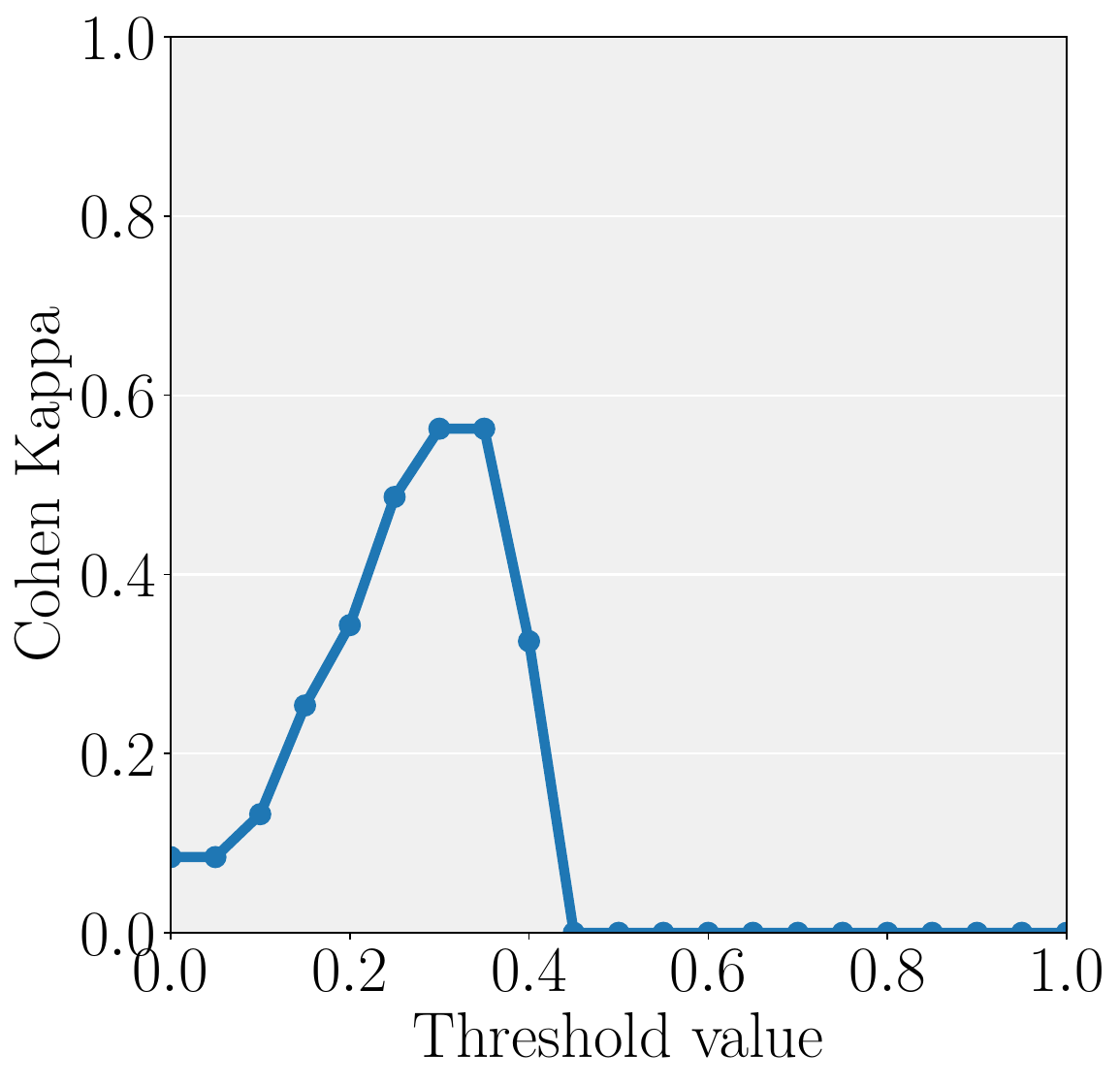}
  \end{minipage}%
  \begin{minipage}{0.28\textwidth}\centering
    \includegraphics[width=0.9\textwidth]{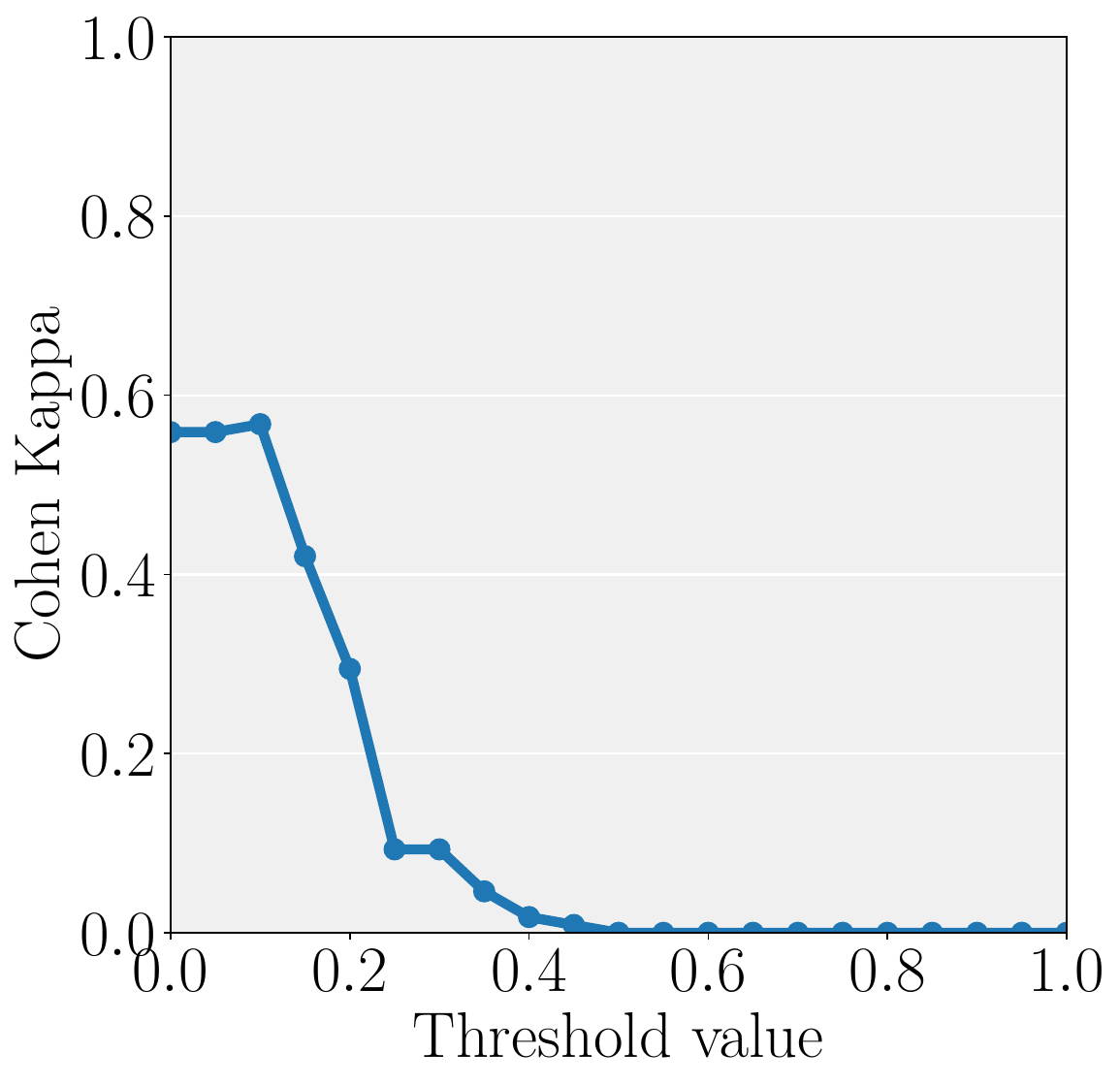}
  \end{minipage}

  \begin{minipage}{0.05\textwidth}
    \centering
    \rotatebox{90}{Rouge-L (Recall)}
  \end{minipage}%
  \begin{minipage}{0.28\textwidth}\centering
    \includegraphics[width=0.9\textwidth]{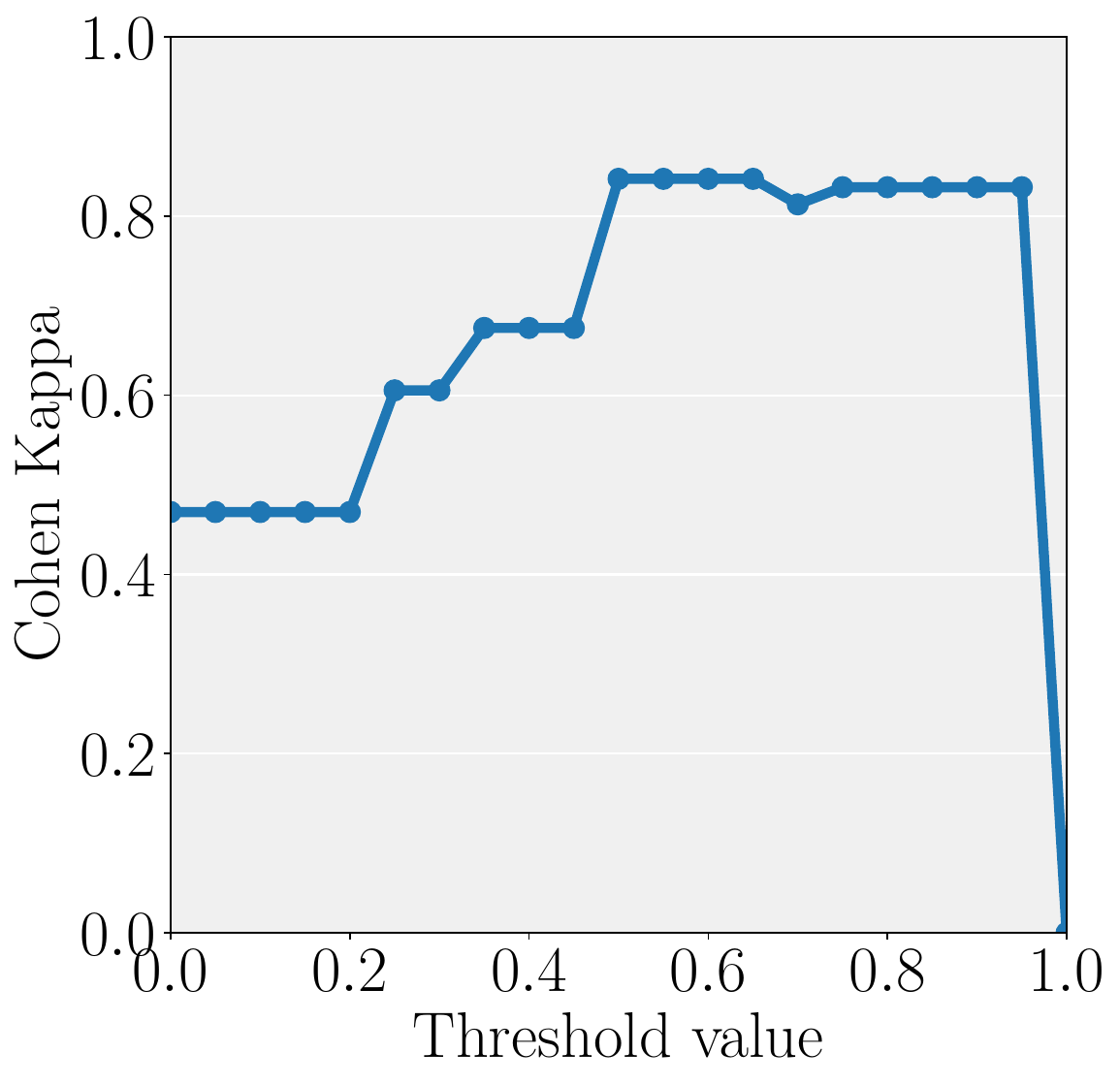}
  \end{minipage}%
  \begin{minipage}{0.28\textwidth}\centering
    \includegraphics[width=0.9\textwidth]{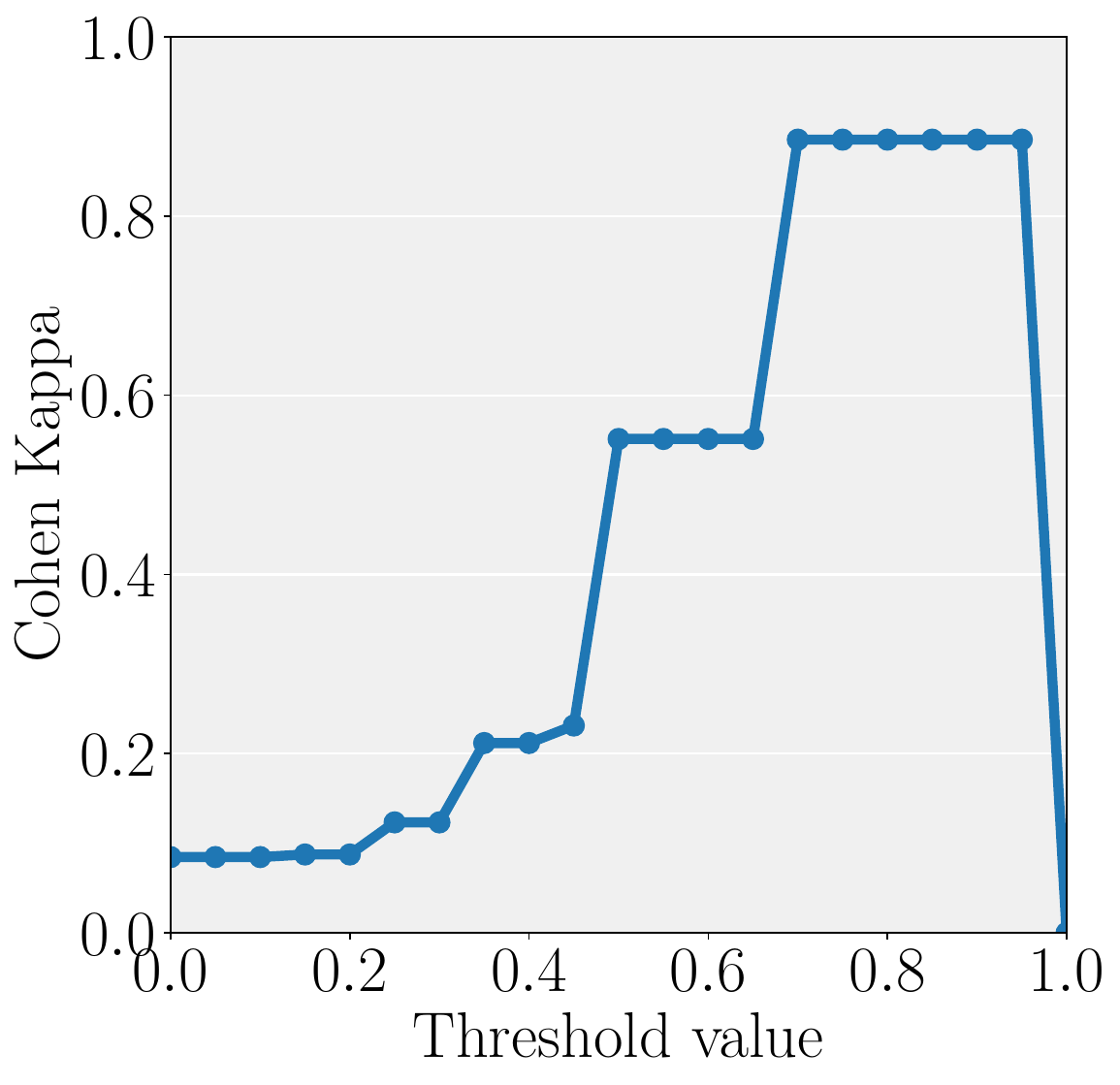}
  \end{minipage}%
  \begin{minipage}{0.28\textwidth}\centering
    \includegraphics[width=0.9\textwidth]{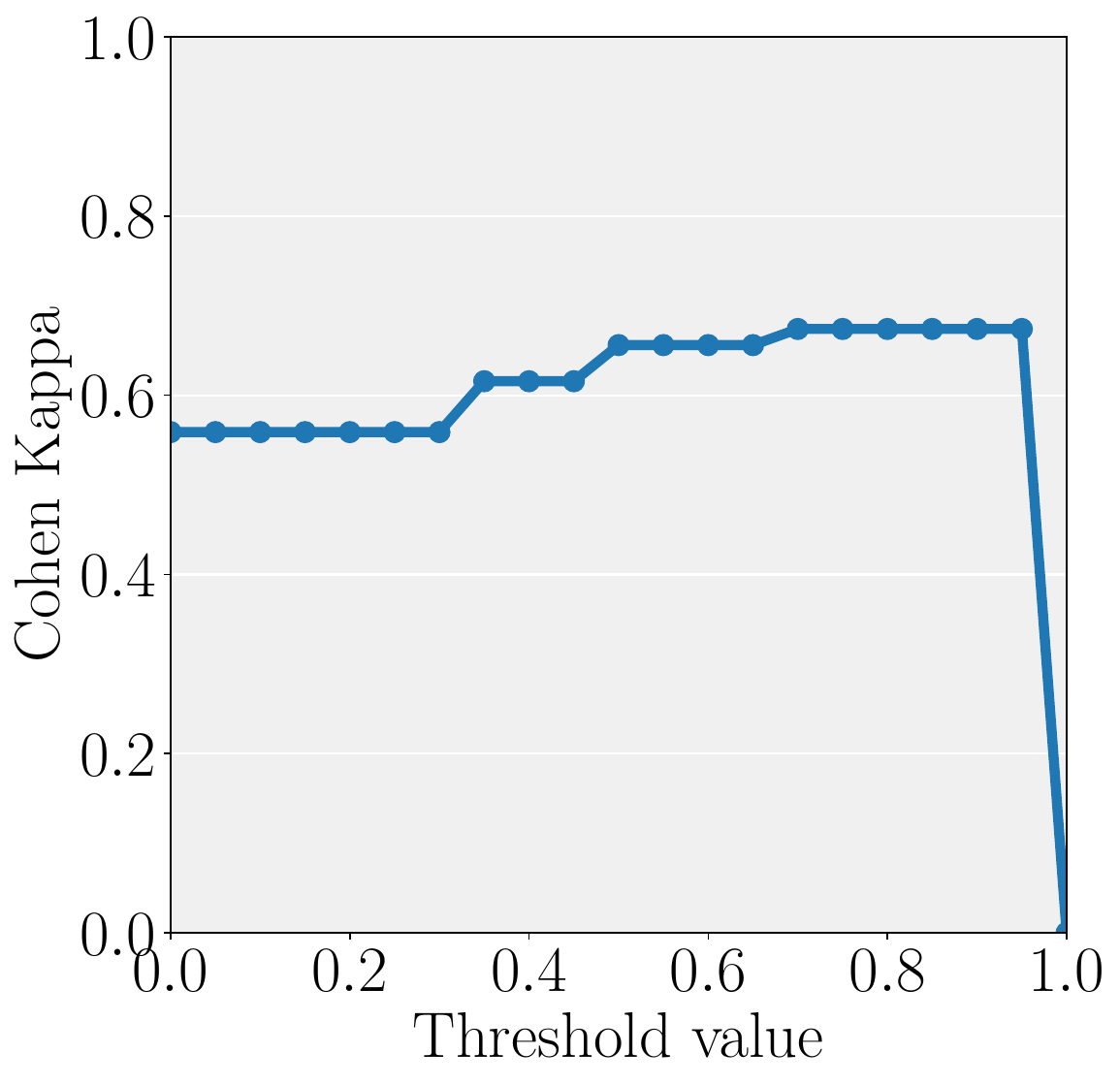}
  \end{minipage}

  \begin{minipage}{0.05\textwidth}
    \centering
    \rotatebox{90}{SentenceBERT}
  \end{minipage}%
  \begin{minipage}{0.28\textwidth}\centering
    \includegraphics[width=0.9\textwidth]{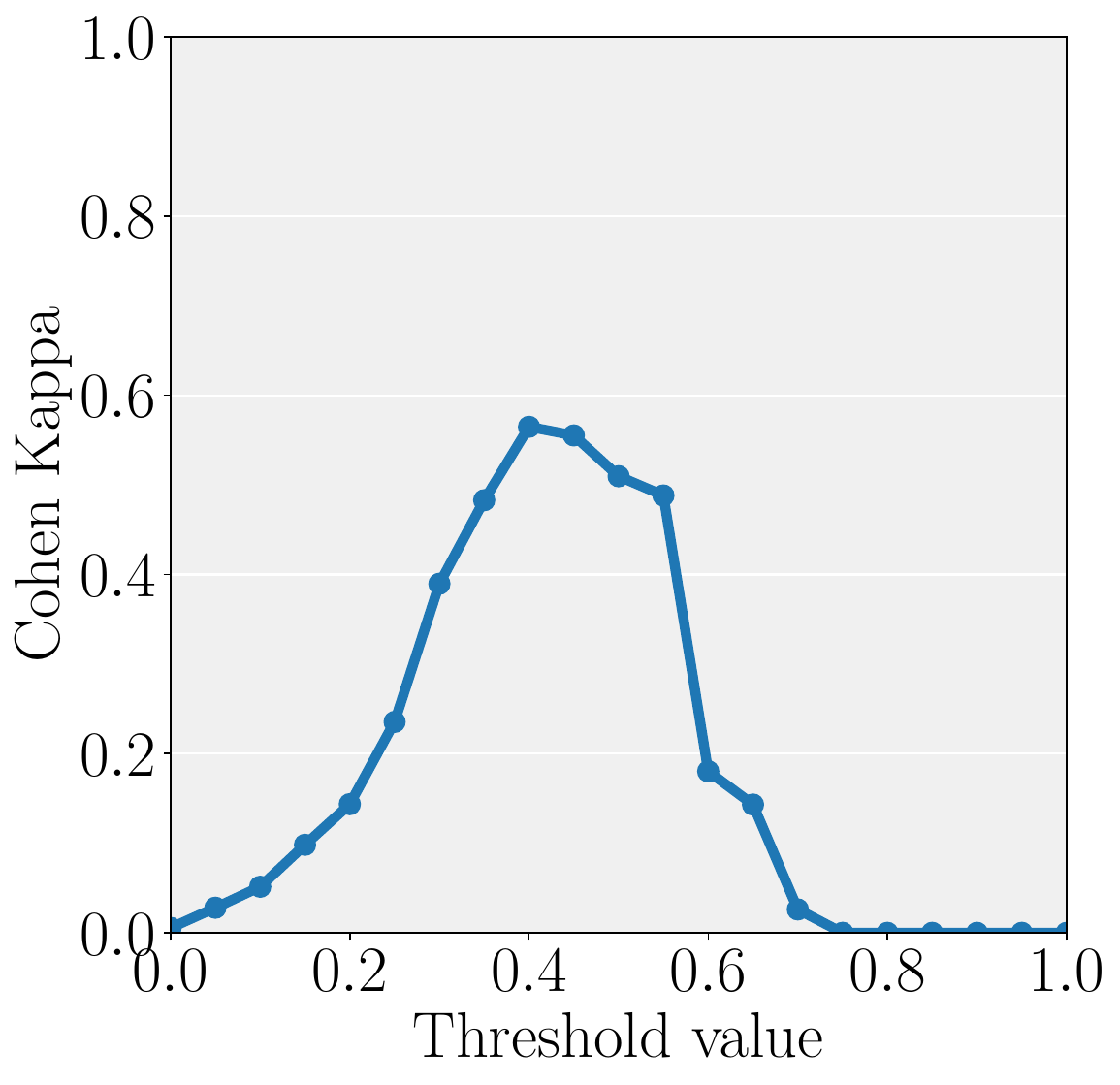}
  \end{minipage}%
  \begin{minipage}{0.28\textwidth}\centering
    \includegraphics[width=0.9\textwidth]{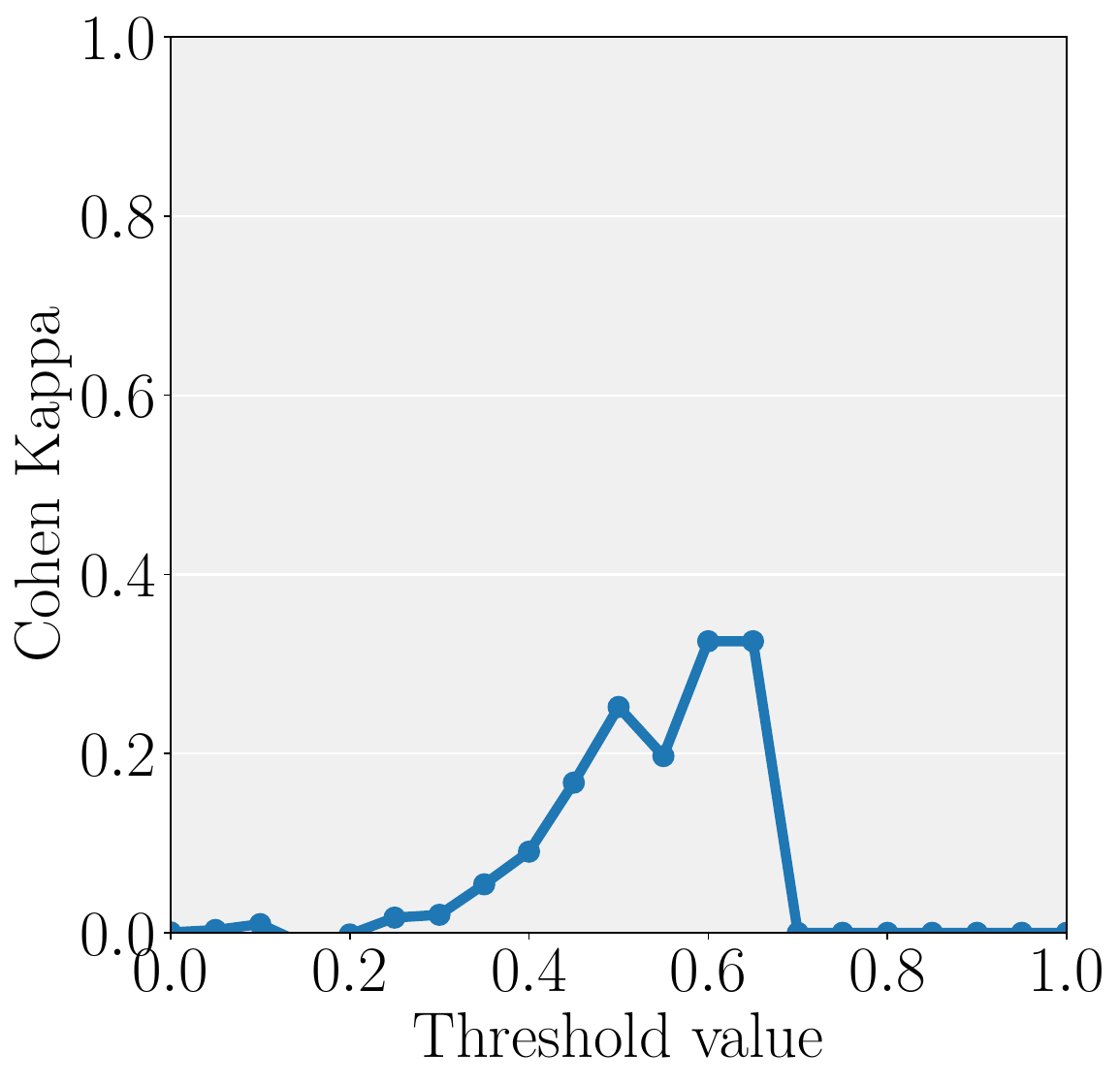}
  \end{minipage}%
  \begin{minipage}{0.28\textwidth}\centering
    \includegraphics[width=0.9\textwidth]{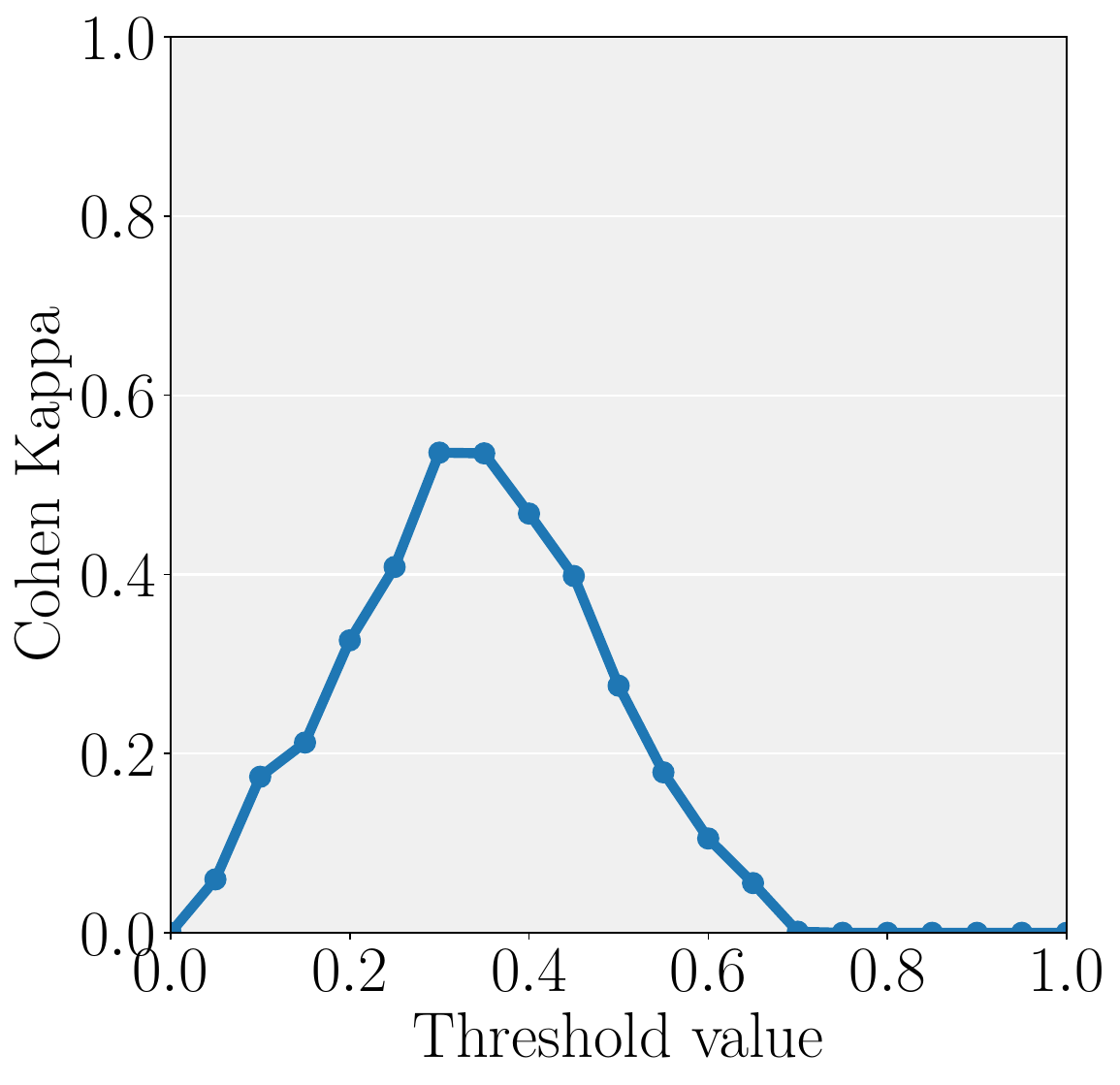}
  \end{minipage}

  \begin{minipage}{0.05\textwidth}
    \centering
    \rotatebox{90}{BERTScore}
  \end{minipage}%
  \begin{minipage}{0.28\textwidth}\centering
    \includegraphics[width=0.9\textwidth]{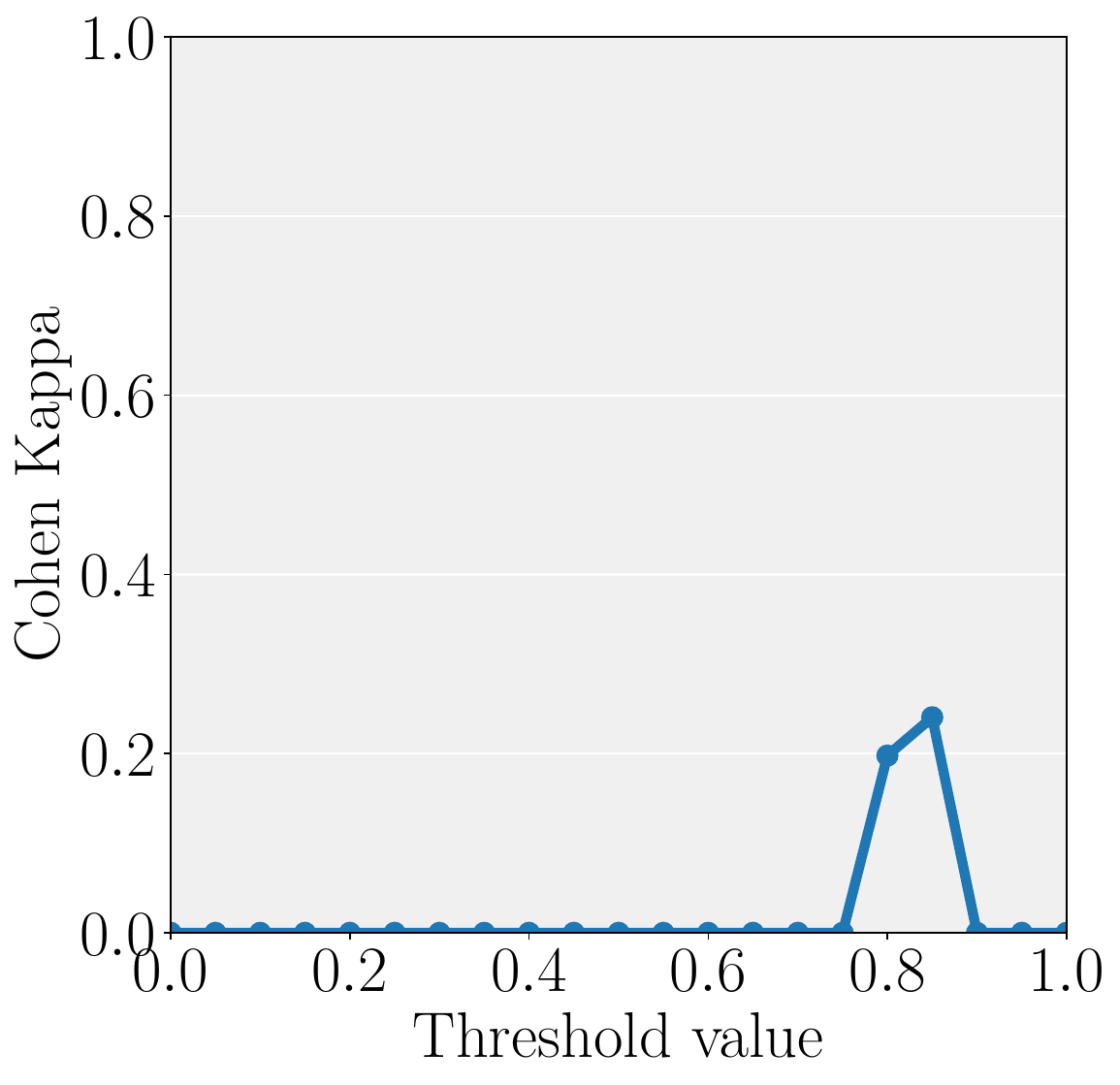}
  \end{minipage}%
  \begin{minipage}{0.28\textwidth}\centering
    \includegraphics[width=0.9\textwidth]{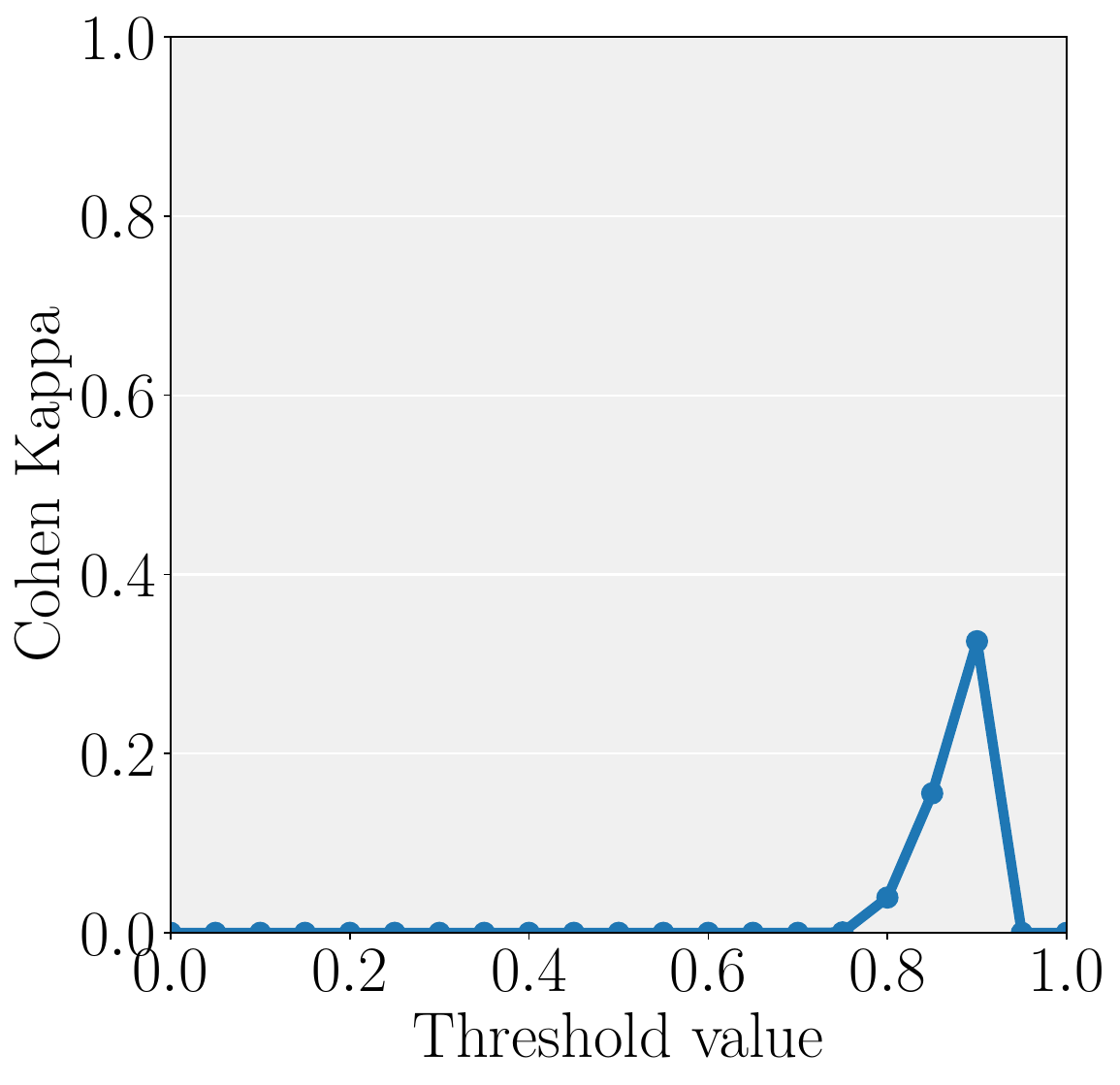}
  \end{minipage}%
  \begin{minipage}{0.28\textwidth}\centering
    \includegraphics[width=0.9\textwidth]{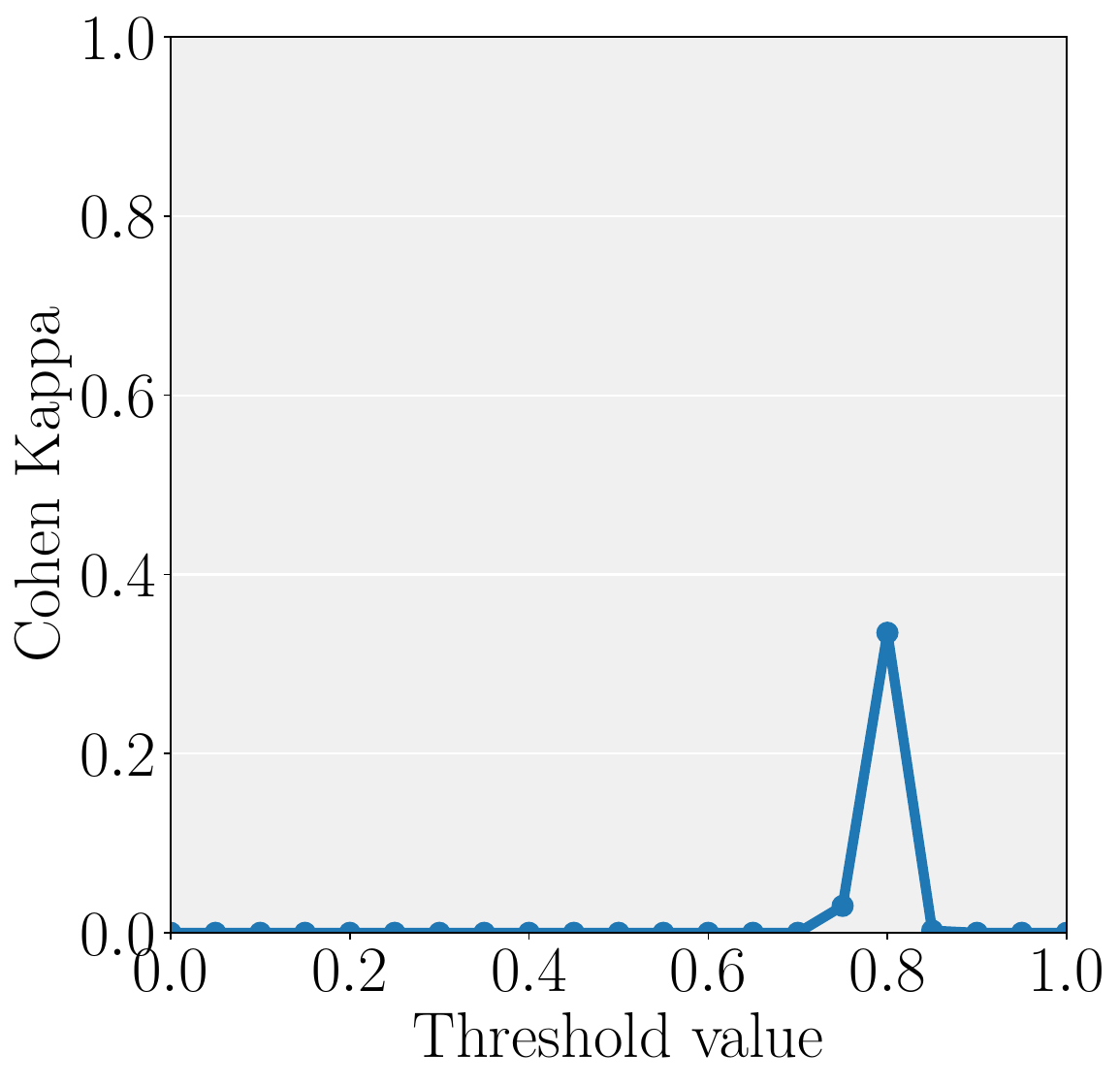}
  \end{minipage}

  \begin{minipage}{0.05\textwidth}
    \centering
    \rotatebox{90}{AlignScore}
  \end{minipage}%
  \begin{minipage}{0.28\textwidth}\centering
    \includegraphics[width=0.9\textwidth]{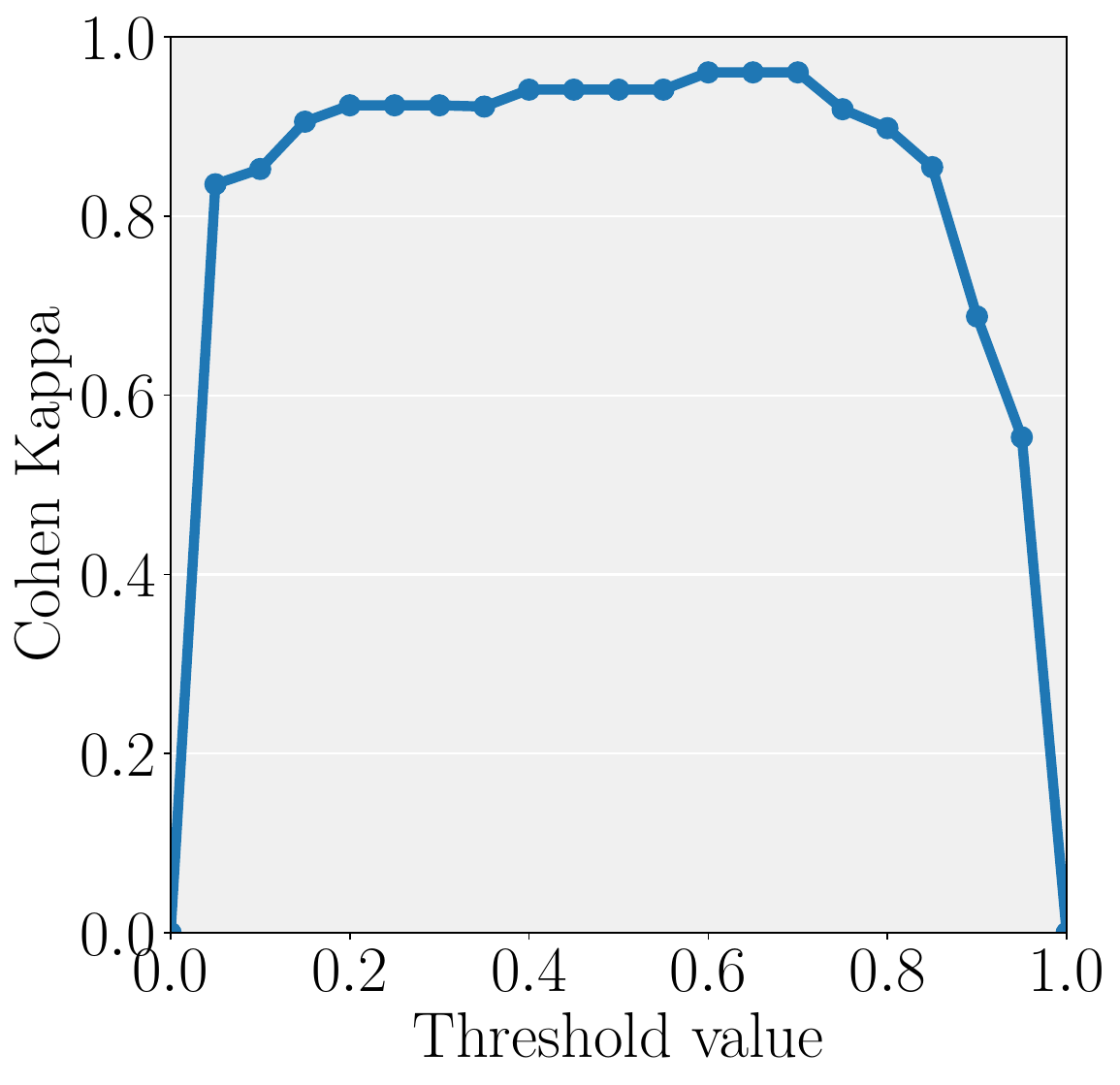}
  \end{minipage}%
  \begin{minipage}{0.28\textwidth}\centering
    \includegraphics[width=0.9\textwidth]{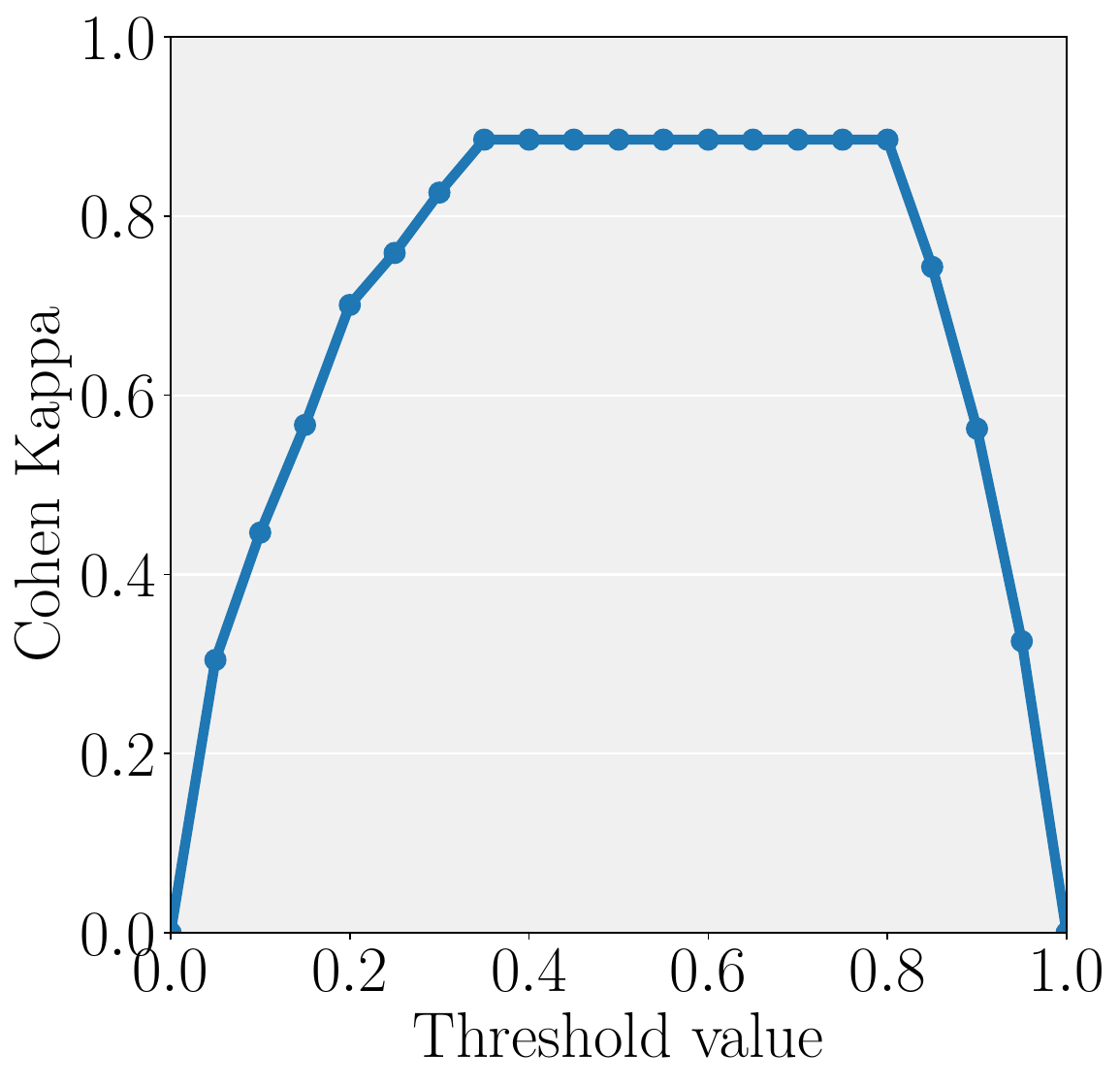}
  \end{minipage}%
  \begin{minipage}{0.28\textwidth}\centering
    \includegraphics[width=0.9\textwidth]{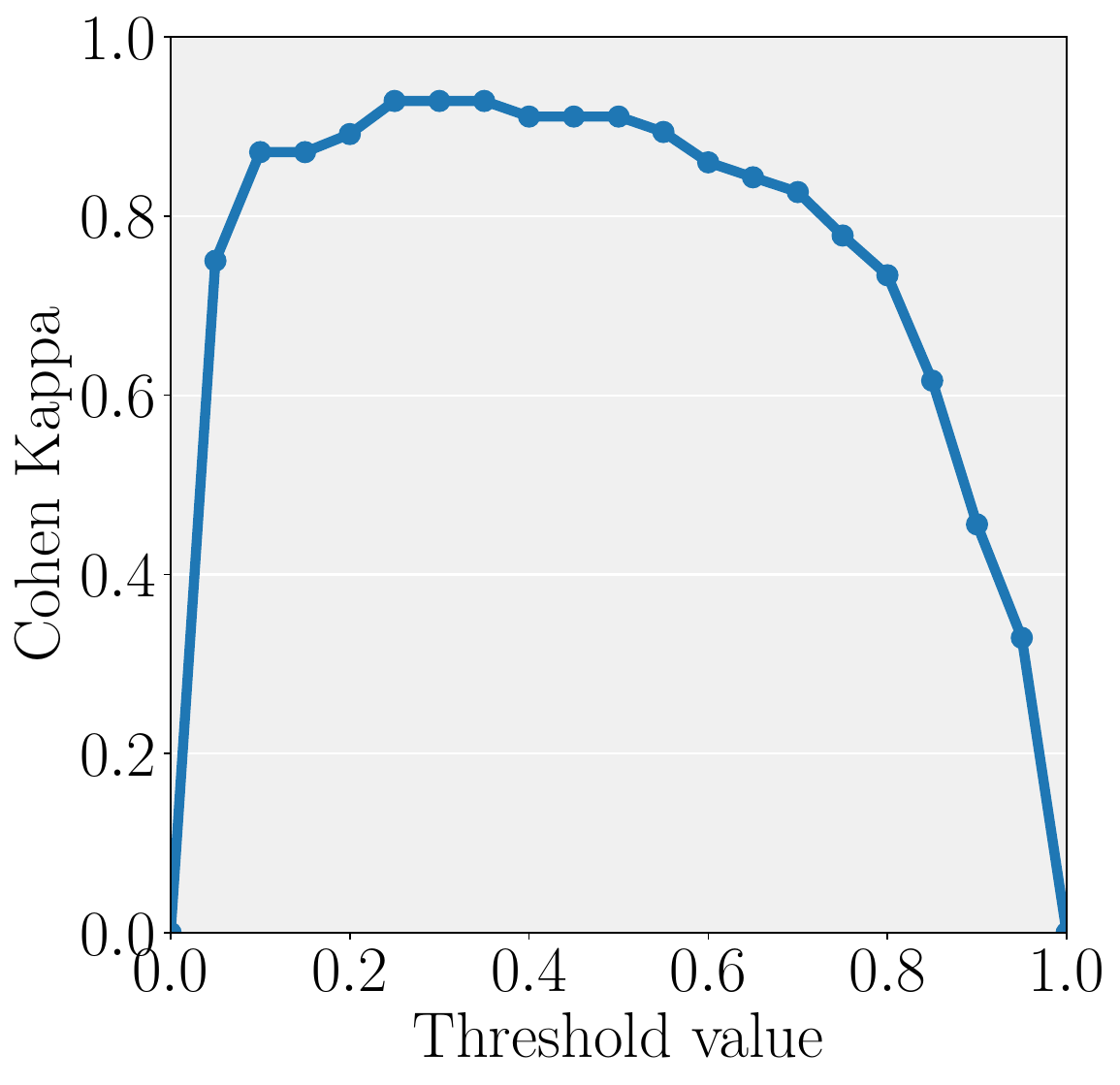}
  \end{minipage}

  \caption{Human‐agreement rate as a function of the \corr threshold.}
  \label{fig:app_threshold_human_agreement}
\end{figure*}

\cref{fig:app_threshold_human_agreement} shows how human-agreement rates vary with the threshold used to binarize different \corrs across datasets. Lexical and embedding-based metrics (e.g., ROUGE-L (F1), BERTScore) show high sensitivity to threshold tuning and inconsistent alignment with human judgments. In contrast, AlignScore yields consistently high agreement across thresholds and datasets.

\end{document}